\documentclass{article}

\usepackage{microtype}
\usepackage{graphicx}
\usepackage{subcaption}
\usepackage{booktabs} 
\usepackage{amsmath}
\usepackage{amsfonts}
\usepackage{hyperref}

\usepackage[accepted]{icml2022}
\usepackage{listings}
\usepackage{inconsolata}
\definecolor{mygreen}{rgb}{0,0.6,0}
\definecolor{mygray}{rgb}{0.5,0.5,0.5}
\definecolor{mymauve}{rgb}{0.58,0,0.82}

\usepackage{amsmath}
\usepackage{amsfonts}
\usepackage{amssymb}
\usepackage{amsthm}
\usepackage{bm}
\usepackage{mathtools}
\usepackage[inline]{enumitem}
\usepackage{thmtools,thm-restate}
\usepackage{algorithm}
\usepackage{algorithmic}
\usepackage{natbib}
\usepackage[capitalise]{cleveref}
\usepackage{xcolor}
\usepackage{graphicx}
\usepackage{comment}

\def\AA{\mathcal{A}}
\def\DD{\mathcal{D}}

\def\MM{\mathcal{M}}

\def\SS{\mathcal{S}}

\def\Ebb{\mathbb{E}}

\def\Pbb{\mathbb{P}}

\usepackage{pifont}
\def\cmark{{\color{green}\ding{51}}}
\def\xmark{{\color{red}\ding{55}}}

\icmltitlerunning{Consistent Dropout}

\begin{document}

\definecolor{commentcolor}{RGB}{110,154,155}   
\newcommand{\PyComment}[1]{\ttfamily\textcolor{commentcolor}{\# #1}}  
\newcommand{\PyCode}[1]{\ttfamily\textcolor{black}{#1}} 

\twocolumn[
\icmltitle{Consistent Dropout for Policy Gradient Reinforcement Learning}




\begin{icmlauthorlist}
\icmlauthor{Matthew Hausknecht}{to}
\icmlauthor{Nolan Wagener}{goo}
\end{icmlauthorlist}

\icmlaffiliation{to}{Microsoft Research}
\icmlaffiliation{goo}{Institute for Robotics and Intelligent Machines, Georgia Institute of Technology}

\icmlcorrespondingauthor{Matthew Hausknecht}{matthew.hausknecht@gmail.com}
\icmlcorrespondingauthor{Nolan Wagener}{nolan.wagener@gatech.edu}

\icmlkeywords{Machine Learning, ICML}

\vskip 0.3in
]



\printAffiliationsAndNotice{}  

\begin{abstract}
Dropout has long been a staple of supervised learning, but is rarely used in reinforcement learning. 
We analyze why na\"ive application of dropout is problematic for policy-gradient learning algorithms and introduce \textit{consistent dropout}, a simple technique to address this instability.
We demonstrate consistent dropout enables stable training with A2C and PPO in both continuous and discrete action environments across a wide range of dropout probabilities.
Finally, we show that consistent dropout enables the online training of complex architectures such as GPT without needing to disable the model's native dropout.
\end{abstract}

\section{Introduction}
\label{sec:intro}

Dropout, the practice of randomly zeroing a fraction of unit activations at select neural network layers, is an established technique for regularizing neural network learning and reducing overfitting \cite{JMLR:v15:srivastava14a}.
However, in the field of reinforcement learning, dropout is rarely applied.
Consider the large body of work without dropout, including high-profile projects like AlphaGo \cite{SilverHuangEtAl16nature}, AlphaStar \cite{alphastarblog}, and OpenAI Five \cite{openai_five} -- none of which mention dropout.
Even models that are commonly supervised-trained with dropout, such as Transformers \cite{vaswani17}, disable dropout entirely when repurposed for reinforcement learning \cite{parisotto19,loynd20}.
This trend is reinforced by our open source tools: Surveying six popular open source RL frameworks\footnote{Surveyed frameworks were Dopamine \cite{castro18dopamine}, RLlib \cite{rllib}, OpenAI Baselines \cite{baselines}, Tianshou \cite{tianshou}, Stable-Baselines3 \cite{stable-baselines3}, and rlpyt \cite{rlpyt}.}, we found none that supported dropout.

We contend that the na\"ive application of dropout in reinforcement learning can be detrimental to learning performance. 
To test this hypothesis, we added dropout layers to the policy networks of A2C and PPO agents in three RL frameworks: RLlib \cite{rllib}, Stable-Baselines3 \cite{stable-baselines3}, and Tianshou \cite{tianshou}.
Agents were then trained on the MuJoCo Half-Cheetah-v3 continuous control environment using the hyperparmeters recommended by their respective codebases.
To summarize results in Appendix \ref{sec:framework_testing}, Tianshou's A2C experienced large negative returns (e.g., $-10^7$) in five of sixteen runs, and all sixteen PPO runs prematurely terminated due to NaN values.
Similarly, RLlib experienced instability in five of six A2C runs and two of six PPO runs.
Stable-Baselines3, on the other hand, completed all runs without error, though we found that A2C experienced large policy loss magnitudes on seven of twelve runs and PPO declined in performance by $21\%$ on average with dropout.
Thus, we conclude that instability stemming from na\"ive application of dropout is not an artifact of a single implementation or codebase but instead affects to various degrees all of the tested frameworks.\footnote{Notably, the use of Tanh nonlinearities in policy networks reduced instability compared to ReLU.
Although we did not perform a detailed investigation, we suspect this relates to Tanh bounding the network outputs.}

To better understand \textit{why} dropout is problematic, \cref{sec:why_dropout} shows the instability stems from different dropout masks being applied to the same observations during rollouts and updates.
This mismatch leads to policy collapse for A2C and reduced performance for PPO.
Furthermore, we show that more complex networks such as GPT \cite{radford2019language} suffer more acutely than MLPs from the instabilities introduced by na\"ive application of dropout.
To correct this instability, \cref{sec:consistent_dropout} introduces \textit{consistent dropout}, a simple and mathematically sound technique that consists of saving the random dropout masks generated during rollouts and reapplying the same masks during updates.
Experimental results in \cref{sec:experiments} show this technique stabilizes learning for both A2C and PPO on discrete and continuous environments across a spectrum of dropout probabilities.
We also show this technique allows for GPT to be trained online without needing to disable dropout.

The contributions of this paper are:
\begin{enumerate*}[label=\arabic*)]
\item Demonstrate that na\"ive application of dropout creates instability for policy gradient methods across a variety of open-source RL frameworks, algorithms, and environments.
\item Analyze the source of instability and introduce a simple method of saving and reapplying dropout masks to stabilize training.
\item Demonstrate stable online RL training of GPT with dropout.
\end{enumerate*}


\section{Related Work}
\label{sec:related}

\citet{farebrother18} investigated dropout as a regularization technique for the DQN learning agent \cite{mnih2015} on Atari environments. 
Their findings indicate that moderate levels of dropout ($p=0.1$), paired with $\ell_2$ regularization improved zero-shot generalization on variants of the training environment.
However, the authors note that dropout slowed the training performance and typically required more training iterations to  compete with the unregularized baseline.

Similarly, \citet{cobbe18} examined the effects of dropout and $\ell_2$ regularization on the Procgen benchmark \cite{cobbe2019procgen}.
Using the PPO algorithm, they found both $\ell_2$ regularization and moderate levels of dropout ($p=0.1$) aided generalization to held-out test environments, but also slightly reduced performance on training environments.\footnote{Inspection of the \href{https://github.com/openai/coinrun}{CoinRun codebase} indicates Cobbe et al. implemented a variant of consistent dropout, but it is neither mentioned in the paper nor analyzed.}

Dropout has been successfully applied in sequential decision making settings when training is based on offline RL or imitation learning.
Examples such as the Decision Transformer \cite{chen21} and Trajectory Transformer \cite{janner21} show that transformer models are capable of learning high quality policies from offline expert data.
In this work we show how such models can be additionally trained with online RL, potentially enabling a paradigm of offline pretraining followed by online finetuning.

\section{Preliminaries}
We consider an infinite-horizon Markov decision process (MDP) defined as the tuple $\MM = (\SS, \AA, p, R, \gamma)$, where $\SS$ is the state space, $\AA$ is the action space, $p(s'|s,a)$ is the transition kernel, $R(s,a)$ is the reward function, and $\gamma$ is the discount factor.
Let $\pi_\theta(a|s)$ be a stochastic policy parameterized by some vector $\theta$.
Given some start state distribution $d_0(s_0)$, the policy and transition kernel induce a trajectory distribution $\rho^{\pi_\theta}(\tau)$, where $\tau = (s_0, a_0, s_1, a_1, \dots)$.
Our goal is to find a parameter setting that maximizes the expected sum of discounted rewards in the MDP:
\[
\max_{\theta} \, J(\theta) = \Ebb_{\rho^{\pi_\theta}(\tau)} \left[ \sum_{t=0}^\infty \gamma^t R(s_t, a_t) \right] .
\]

We define the value function $V^{\pi_\theta}(s)$ and state-action value function $Q^{\pi_\theta}(s,a)$ as:
\begin{align*}
V^{\pi_\theta}(s) &= \Ebb_{\rho^{\pi_\theta}(\tau)} \left[ \sum_{t=0}^\infty \gamma^t R(s_t, a_t) \;\Bigg|\; s_0 = s \right] \\
Q^{\pi_\theta}(s,a) &= \Ebb_{\rho^{\pi_\theta}(\tau)} \left[ \sum_{t=0}^\infty \gamma^t R(s_t, a_t) \;\Bigg|\; s_0 = s, a_0 = a \right]
\end{align*}
and furthermore define the advantage function ${A^{\pi_\theta}(s,a) = Q^{\pi_\theta}(s,a) - V^{\pi_\theta}(s)}$.

Let $d_t^{\pi_\theta}(s)$ be the state distribution induced by running $\pi_\theta$ in $\MM$ for $t$ time steps from the initial state distribution $d_0$.
Defining the state visitation ${d^{\pi_\theta}(s) = (1-\gamma) \sum_{t=0}^\infty \gamma^t d_t^{\pi_\theta}(s)}$, it can be shown \cite{sutton2000policy} that the policy gradient has the form
\begin{equation}
\label{eq:policy gradient}
\nabla J(\theta) = \Ebb_{d^{\pi_\theta}(s)} \Ebb_{\pi_\theta(a|s)} \left[ A^{\pi_\theta}(s,a) \nabla_\theta \log \pi_\theta (a|s) \right] .
\end{equation}

Practical implementations of policy gradient algorithms roll out $\pi_\theta$ and collect state-action-reward triplets $(s, a, r)$ into a buffer $\DD$.
After learning an approximate advantage function $\hat{A}(s, a)$ (e.g., through generalized advantage estimation~\cite{schulman2016high}), they form the loss function
\begin{equation} \label{eq:policy gradient loss}
L_{PG}(\theta) = -\Ebb_{(s,a)\sim\DD}[ \hat{A}(s, a) \log \pi_\theta(a|s) ]
\end{equation}
and subsequently perform a gradient descent step.

\section{Why Dropout Disrupts Policy Gradient}
\label{sec:why_dropout}
\begin{figure}[t]
    \centering
    \includegraphics[width=\linewidth]{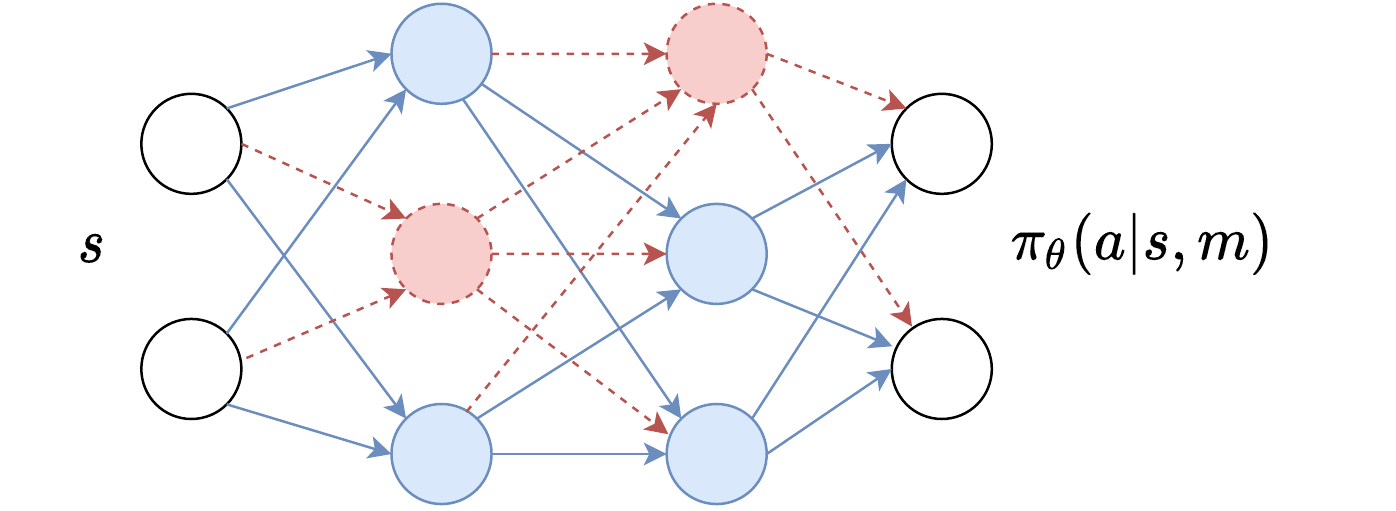}
    \caption{An example of a policy after a dropout mask $m$ has been sampled. The neurons in red have their activations zeroed out.}
    \label{fig:dropout}
\end{figure}

We now assume that $\pi_\theta$ is a multi-layer neural network where there is a subset of neurons wherein each activation is zeroed out with some probability, as shown in \cref{fig:dropout}.
Letting $m$ be the resulting dropout mask, our stochastic policy has the form $\pi_\theta(a|s) = \sum_m p(m) \pi_\theta(a|s,m)$, where $p(m)$ is the probability function of the mask.
In practice, dropout is treated as an unobserved internal process of the network, so that we observe a noisy estimator of $\pi_\theta(a|s)$.

Due to this practice, at rollout time, we sample some dropout mask $\hat{m}$ at a given state $s$ and then sample an action from $\pi_\theta(a|s,\hat{m})$ to apply in the MDP.
At update time, we sample a new dropout mask when querying the policy for its log-probability.
This results in a modified loss function for policy optimization:
\[
\widetilde{L}_{PG}(\theta) = -\Ebb_{(s,a)\sim\DD} \Ebb_{p(m)} [ \hat{A}(s, a) \log \pi_\theta(a|s,m) ] .
\]
The use of different dropout masks for the same state $s$ at rollout time versus update time can lead to vastly different action probabilities.
For A2C, this manifests in $\pi_\theta(a|s)$ being close to zero and $\log\pi_\theta(a|s)$ approaching negative infinity.
At this point, updates become unstable as policy loss and gradient magnitudes explode.
In practice, gradient norm clipping can keep training numerically stable, but learning performance still suffers due to the gradient estimator having high variance.
Table \ref{tab:example} demonstrates how the application of different dropout masks can change the outputs and lower the log probabilities of actions in an untrained, newly initialized network.
These effects only become more pronounced over the course of training.

\begin{table}[t]
\caption{\textbf{Forward passes over the same state with different dropout masks change output actions and reduce the log probabilities:}
Let $a^0$ and $a^1$ be the most likely (i.e., mode) actions from two forward passes on the same state $s$ using random dropout masks $m^0$ and $m^1$, respectively.
For continuous actions, $d(a^0, a^1)$ is the absolute difference $|a^0 - a^1|$.
For discrete actions, $d(a^0, a^1)$ indicates the frequency of disagreement $\Pbb[a^0 \neq a^1]$.
Higher levels of dropout result in greater differences in actions as well as lower log probabilities. This effect is more pronounced in GPT, in which a dropout probability of $0.1$ is approximately equivalent to an MLP with dropout of $0.75$.}
\label{tab:example}
\vskip 0.15in
\begin{center}
\begin{small}
\begin{sc}
\setlength{\tabcolsep}{4pt}
\begin{tabular}{lccc}
\toprule
Net & Dropout & $d(a^0, a^1)$ & $\log \pi_\theta(a^0 | s, m^1)$ \\
\midrule
3-Layer & $0$ & $0 \pm 0$ & $-0.83 \pm 0.00$ \\
MLP w/ & $0.1$ & $0.04 \pm 0.03$ & $-0.84 \pm 0.01$ \\
4 continuous & $0.25$ & $0.06 \pm 0.05$ & $-0.86 \pm 0.02$ \\
actions & $0.5$ & $0.12 \pm 0.09$ & $-0.90 \pm 0.07$ \\
 & $0.75$ & $0.23 \pm 0.25$ & $-1.13 \pm 0.38$ \\
 & $0.9$ & $0.63 \pm 0.65$ & $-6.36 \pm 7.04$ \\
\hline
4-Layer & $0$ & $0 \pm 0$ & $-0.83 \pm 0.00$ \\
GPT w/ & $0.1$ & $0.26 \pm 0.20$ & $-1.14 \pm 0.33$ \\
4 continuous & $0.25$ & $0.41 \pm 0.32$ & $-1.57 \pm 0.77$ \\
actions & $0.5$ & $0.52 \pm 0.40$ & $-2.01 \pm 1.21$ \\
 & $0.75$ & $0.62 \pm 0.47$ & $-2.47 \pm 1.63$ \\
 & $0.9$ & $0.64 \pm 0.49$ & $-2.64 \pm 1.80$ \\
\hline
3-Layer & $0$ & $0 \pm 0$ & $-1.31 \pm 0.03$ \\
MLP w/  & $0.1$ & $0.18 \pm 0.39$ & $-1.35 \pm 0.04$ \\
4 discrete & $0.25$ & $0.51 \pm 0.50$ & $-1.34 \pm 0.05$ \\
actions & $0.5$ & $0.60 \pm 0.48$ & $-1.34 \pm 0.10$ \\
& $0.75$ & $0.71 \pm 0.47$ & $-1.39 \pm 0.21$ \\
 & $0.9$ & $0.75 \pm 0.43$ & $-1.51 \pm 0.58$ \\
\bottomrule
\end{tabular}
\end{sc}
\end{small}
\end{center}
\vskip -0.1in
\end{table}

\subsection{Dropout and PPO}
PPO \cite{schulman17} optimizes a clipped objective that ensures the deviation of the current policy $\pi_\theta$ from the rollout policy $\pi_{\theta_\mathrm{old}}$ is small:
{\small
\begin{align*}
L_{PPO}(\theta) = -\Ebb_\DD[ \min\{& r(s,a,\theta) \hat{A}(s,a), \\
                                   & \mathrm{clip}(r(s,a,\theta), 1-\epsilon, 1+\epsilon) \hat{A}(s,a) \} ],
\end{align*}}%
where $r(s, a, \theta) = \frac{\pi_\theta(a|s)}{\pi_{\theta_\mathrm{old}}(a|s)}$ is the ratio of the action probabilities under the new and old policies, respectively, and $\epsilon$ is a small constant which determines how far the policy is allowed to diverge from the old policy. 
In addition to this clipped objective, PPO also has the option to stop taking gradient steps if the KL divergence between the new and old policies grows past a threshold.
This early stopping mechanism can help ensure monotonic policy improvement by preventing gradient updates from extrapolating too far in any direction.

Adding dropout to the policy network increases the variance in the ratio of action probabilities $r$. 
The PPO objective clips this variance and prevents policy loss $L_{PPO}(\theta)$ from exploding.
When using an early stopping threshold, we find that increased variance in $r$ will often trigger immediate early stopping, preventing any updates from taking place.
To preface our results, without early stopping we observe stable learning but notably reduced performance.

\section{Policy Gradient with Dropout}
\label{sec:consistent_dropout}

We now present two methods to correctly compute the policy gradient in \cref{eq:policy gradient} in the presence of dropout.

\subsection{Dropout-Marginalized Gradient} \label{sec:dropout-marginalized gradient}

This method gives the correct policy gradient by marginalizing over possible dropout masks that are randomly sampled at update time:
For some $s \in \SS$ and $a \in \AA$, the policy's score function can be computed as:
\begin{align}
\nabla_\theta \log \pi_\theta(a|s) &= \nabla_\theta \log\left( \sum_m p(m) \pi_\theta(a|s,m) \right) \nonumber \\
&= \frac{\sum_m p(m) \pi_\theta(a|s,m) \nabla_\theta \log\pi_\theta(a|s,m)}{\sum_m p(m) \pi_\theta(a|s,m)} \label{eq:marginalized gradient} \\
&= \sum_m p(m|s,a,\theta) \nabla_\theta \log \pi_\theta(a|s,m) \nonumber .
\end{align}
Thus, the score function is the expected score function of the dropout-conditioned policy under the posterior distribution of the dropout mask.

In this work, we use \cref{eq:marginalized gradient} to compute the score function.
Since it is intractable to enumerate every dropout mask, we  instead rely on sampling.
For some $s \in \SS$ and $a \in \AA$, we sample $N$ dropout masks $m_1, \dots, m_N$ i.i.d. from $p(m)$.
For some mask $m_n$, we compute the weight
\[
w_n = \frac{p(m_n) \pi_\theta(a|s,m_n)}{\sum_{k=1}^N p(m_k) \pi_\theta(a|s,m_k)},
\]
whereupon the score function can be estimated as:
\[
\nabla_\theta \log\pi_\theta(a|s) \approx \sum_{n=1}^N w_n \nabla_\theta \log\pi_\theta(a|s,m_n) .
\]
We note that the weights approximate the mask posterior, i.e., $w_n \approx p(m_n | s, a, \theta)$.
The estimated score function can then be used to compute the policy gradient in \cref{eq:policy gradient}.

This method has the advantage of not needing to store the dropout mask $m$ when rolling out the policy $\pi_\theta$, but the multiple samples required to compute the approximate mask posterior can lead to computational overhead for larger networks.

\subsection{Dropout-Conditioned Gradient} \label{sec:dropout-conditioned gradient}

\begin{figure}
    \centering
    \includegraphics[width=0.5\textwidth]{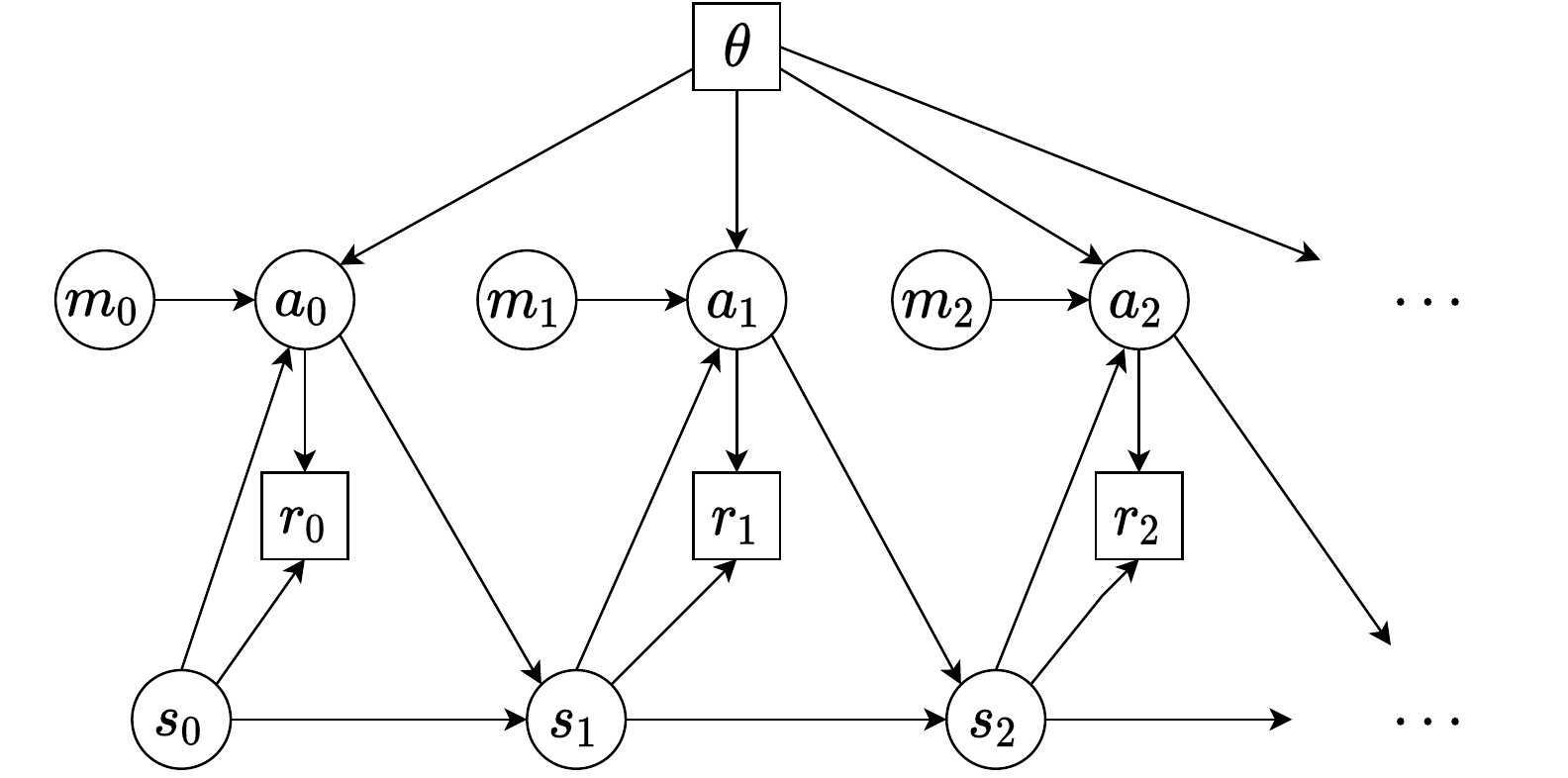}
    \caption{Stochastic computation graph \cite{schulman2015gradient} of the MDP including the dropout mask $m$}
    \label{fig:bayes net}
\end{figure}

Consider the modified MDP (\cref{fig:bayes net}) in which the policy outputs both the action and the dropout mask $(a,m)$, so that the stochastic policy is now ${\pi_\theta(a,m|s) = p(m) \pi_\theta(a|s,m)}$.
This MDP is equivalent to the original since neither the dynamics nor the reward are modified, i.e., ${p(s'|s,a,m) = p(s'|s,a)}$ and ${R(s,a,m) = R(s,a)}$.
As a consequence, the advantage function $A^{\pi_\theta}$ remains a function only of the state $s$ and original action $a$.

Now, the score function of the policy is:
\begin{align*}
\nabla_\theta \log\pi_\theta(a,m|s) &= \nabla_\theta(\log p(m) + \log \pi_\theta(a|s,m)) \\
&= \nabla_\theta \log\pi_\theta(a|s,m)
\end{align*}%
so that the resultant policy gradient is
\begin{align*}
\nabla J(\theta) &= \Ebb_{d^{\pi_\theta}(s)} \Ebb_{\pi_\theta(a,m|s)}\left[ A^{\pi_\theta}(s,a) \nabla_\theta \log\pi_\theta(a,m|s) \right] \\
&= \Ebb_{d^{\pi_\theta}(s)} \Ebb_{\pi_\theta(a,m|s)}\left[ A^{\pi_\theta}(s,a) \nabla_\theta \log\pi_\theta(a|s,m) \right] .
\end{align*}

Thus, we conclude that conditioning the policy on the same mask $m$ that is sampled during rollouts results in an \mbox{unbiased} policy gradient at update time.
Finally, we note that while the derivations in the preceding sections are specialized to dropout, they can be easily extended to other networks that internally use stochasticity, such as Bayesian neural networks and hidden layers that rely on sampling.

\subsection{Implementation}

From an implementation perspective, due to the computational challenges of using the dropout-marginalized gradient (explored in \cref{sec:continuous action environments}), we focus on the dropout-conditioned gradient for the remainder of this work and refer to it as \emph{consistent dropout}.
At rollout time, for some state $s$, we sample a mask $m$ and action $a$ from $\pi_\theta(a,m|s)$, observe a reward $r$, and store the tuple $(s,a,r,m)$ into a buffer $\DD$.
During subsequent updates, the saved masks can be reapplied with the corresponding states and actions.
The resulting policy loss we optimize has the form:
\[
L_{PG}(\theta) = -\Ebb_{(s,a,m) \sim \DD} [\hat{A}(s,a) \log \pi_\theta(a|s,m)] .
\]
This reuse of dropout masks ensures that any change in the policy network's output is due to a change in the policy network's weights, rather than a different dropout mask.

To facilitate easy storage and retrieval of masks, we design a custom dropout layer shown in Figure \ref{fig:cdropout_layer} that retains the most recent masks from the forward pass, where they can be easily stored and reapplied at update time.
From a computational perspective, this approach is essentially just as fast as the original, but does require additional storage of dropout masks whose size and number vary based on the architecture of the policy net. 
However, by storing masks as binary arrays, we find they often take less space than states or actions.
Finally, additional source code and examples showing the use of this consistent dropout layer may be found in the supplementary materials.

\begin{figure*}[htp]
\vskip 0.2in
\begin{center}
\begin{lstlisting}[language=Python]
import torch
from torch import nn

class ConsistentDropout(nn.Module):
    """
    Consistent Dropout Layer uses masks provided in source and stores masks in sink.
    """
    def __init__(self, source, sink, p=0.5):
        super().__init__()
        self.source = source
        self.sink = sink
        self.p = p
        self.scale_factor = 1 / (1 - self.p)
        
    def forward(self, x):
        if self.training:
            if self.source:
                mask = self.source.pop(0)
            else:
                mask = torch.bernoulli(x.data.new(x.data.size()).fill_(1-self.p))
            self.sink.append(mask.bool())
            return x * mask * self.scale_factor
        else:
            return x


            
class ConsistentPolicyNet(nn.Module):
    def __init__(self, obs_size, action_size, hidden_size, dropout):
        super().__init__()
        self.source, self.sink = [], []
        self.policy_net = nn.Sequential(
            nn.Linear(obs_size, hidden_size),
            nn.ReLU(),
            ConsistentDropout(self.source, self.sink, p=dropout),
            nn.Linear(hidden_size, hidden_size),
            nn.ReLU(),
            ConsistentDropout(self.source, self.sink, p=dropout),
            nn.Linear(hidden_size, action_size)
        )
        
    def forward(self, obs, dropout_masks=None):
        if dropout_masks:
            self.source.extend(dropout_masks)
        action = self.policy_net(obs)
        masks = self.sink.copy()
        self.source.clear()
        self.sink.clear()
        return action, masks
\end{lstlisting}
\caption{\textbf{Consistent Dropout Layer}: PyTorch implementation of the consistent dropout layer and modifications needed to integrate into a policy network: The \texttt{ConsistentDropout} layer will retrieve and apply a dropout mask if provided in \texttt{source} and save the used dropout mask to \texttt{sink}.  \texttt{ConsistentPolicyNet} demonstrates the integration of the \texttt{ConsistentDropout} layer into a policy network: Both \texttt{source} and \texttt{sink} are implemented using lists shared between all dropout layers in the network. In the forward pass, if dropout masks are provided they will be applied in correct order. Otherwise, random masks will be generated and returned. Returned masks can then be stored in the agent's replay memory  and reapplied at update time.}
\label{fig:cdropout_layer}
\end{center}
\vskip -0.2in
\end{figure*}

\section{Experiments}
\label{sec:experiments}

\begin{table}[htb]
\caption{\textbf{Relative Performance Under Dropout:} Training performance relative to a baseline that does not use dropout. Across most environments, A2C with inconsistent dropout becomes unstable and achieves nearly zero performance. Consistent A2C (A2C-C) and Consistent PPO (PPO-C) achieve significantly more performance than their inconsistent counterparts on all MuJoCo environments. In Atari environments, PPO and PPO-C are much closer in relative performance with the trend favoring PPO-C.}
\label{tab:results}
\vskip 0.15in
\begin{center}
\begin{scriptsize}
\begin{sc}
\begin{tabular}{c||cc|cc}
\toprule
Drop & A2C & A2C-C & PPO & PPO-C \\
\hline \hline
\multicolumn{5}{c}{\textbf{HalfCheetah-v3}} \\
$0.1$ & $0 \pm 0$ & $\mathbf{0.77} \pm 0.03$ & $0.61 \pm 0.08$ & $\mathbf{0.95} \pm 0.07$ \\
$0.25$ & $0 \pm 0$ & $\mathbf{0.72} \pm 0.09$ & $0.21 \pm 0.01$ & $\mathbf{0.82} \pm 0.12$ \\
$0.5$ & $0 \pm 0$ & $\mathbf{0.49} \pm 0.03$ & $0.07 \pm 0.09$ & $\mathbf{0.55} \pm 0.11$ \\
\hline
\multicolumn{5}{c}{\textbf{Hopper-v3}} \\
$0.1$ & $0 \pm 0$ & $\mathbf{1.04} \pm 0.06$ & $\mathbf{1.08} \pm 0.31$ & $1.04 \pm 0.2$\\
$0.25$ & $0 \pm 0$ & $\mathbf{1.05} \pm 0.03$ & $0.79 \pm 0.10$ & $\mathbf{1.00} \pm 0.21$\\
$0.5$ & $0 \pm 0$ & $\mathbf{0.98} \pm 0.05$ & $0.31 \pm 0.05$ & $\mathbf{1.08} \pm 0.34$\\
\hline
\multicolumn{5}{c}{\textbf{Walker2d-v3}} \\
$0.1$ & $0 \pm 0$ & $\mathbf{0.90} \pm 0.10$ & $0.27 \pm 0.05$ & $\mathbf{0.91} \pm 0.08$\\
$0.25$ & $0 \pm 0$ & $\mathbf{0.96} \pm 0.14$ & $0.08 \pm 0.01$ & $\mathbf{0.55} \pm 0.05$\\
$0.5$ & $0 \pm 0$ & $\mathbf{0.50} \pm 0.08$ & $0.02 \pm 0.00$ & $\mathbf{0.40} \pm 0.03$\\
\hline
\multicolumn{5}{c}{\textbf{Beamrider}} \\
$0.1$ & $0.14 \pm 0.10$ & $\mathbf{0.48} \pm 0.07$ & $0.38 \pm 0.04$ & $\mathbf{0.70} \pm 0.22$\\
$0.25$ & $0.08 \pm 0.06$ & $\mathbf{0.43} \pm 0.03$ & $0.40 \pm 0.07$ & $\mathbf{0.51} \pm 0.25$\\
$0.5$ & $0.07 \pm 0.10$ & $\mathbf{0.23} \pm 0.09$ & $0.16 \pm 0.02$ & $\mathbf{0.17} \pm 0.03$\\
\hline
\multicolumn{5}{c}{\textbf{Breakout}} \\
$0.1$ & $0 \pm 0$ & $\mathbf{1.12} \pm 0.63$ & $0.64 \pm 0.03$ & $\mathbf{0.77} \pm 0.04$\\
$0.25$ & $0.01 \pm 0$ & $\mathbf{0.94} \pm 0.13$ & $0.56 \pm 0.10$ & $\mathbf{0.68} \pm 0.08$\\
$0.5$ & $0.01 \pm 0$ & $\mathbf{0.36} \pm 0.13$ & $\mathbf{0.53} \pm 0.08$ & $0.28 \pm 0.03$\\
\hline
\multicolumn{5}{c}{\textbf{Seaquest}} \\
$0.1$ & $1.56 \pm 1.45$ & $\mathbf{2.08} \pm 0.53$ & $\mathbf{1.13} \pm 0.09$ & $1.10 \pm 0.25$\\
$0.25$ & $0.34 \pm 0.39$ & $\mathbf{1.57} \pm 0.34$ & $\mathbf{0.99} \pm 0.06$ & $0.97 \pm 0.07$\\
$0.5$ & $1.5 \pm 1.33$ & $\mathbf{2.13} \pm 0.89$ & $1.01 \pm 0.12$ & $\mathbf{1.10} \pm 0.32$\\
\bottomrule
\end{tabular}
\end{sc}
\end{scriptsize}
\end{center}
\vskip -0.1in
\end{table}

In order to compare consistent and inconsistent dropout, we evaluate A2C and PPO on several popular continuous and discrete environments.
In particular, the aim of our experiments is to empirically understand the effects of inconsistent dropout on policy gradient algorithms and to understand if consistent dropout stabilizes policy gradient learning.

\subsection{Continuous-Action Environments} \label{sec:continuous action environments}
We conduct experiments on the Hopper, Walker2d, and Half-Cheetah MuJoCo tasks to compare the stability and performance of A2C and PPO when trained with consistent and inconsistent dropout probabilities of $(0/ 0.1/ 0.25/ 0.5)$.

Results on Half-Cheetah are shown in Figure \ref{fig:half_cheetah} and similar results from the other MuJoCo tasks may be found in Appendix \ref{sec:full_mujoco_results}.
Across all MuJoCo environments, we find that A2C is highly unstable for any level of inconsistent dropout.
PPO is more robust to inconsistent dropout and often remains stable and performant at dropout of $0.1$. 
However, dropout probabilities of $0.25$ and $0.5$ degrade PPO's performance significantly.

On the other hand, as shown in \cref{tab:results}, consistent dropout effectively stabilizes A2C and PPO across all MuJoCo environments and dropout levels.
All levels of consistent dropout lead to stable learning, although dropout of $0.5$ notably reduces performance for both algorithms.

In terms of absolute performance, baseline models without dropout perform slightly better on average than equivalent models trained with dropout.
This finding reflects the intuition that when training and testing on the same environment, less regularization is likely beneficial.

\begin{figure}[htb]
\begin{center}
\begin{subfigure}{.23\textwidth}
    \centering
    \includegraphics[width=\linewidth]{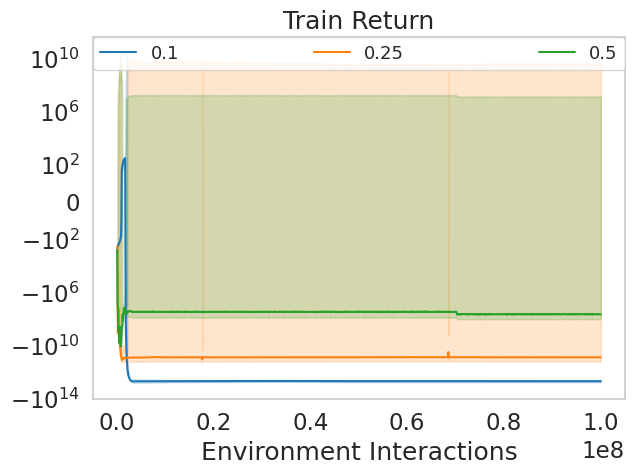}
    \subcaption{$N=10$}
\end{subfigure}
\begin{subfigure}{.23\textwidth}
    \centering
    \includegraphics[width=\linewidth]{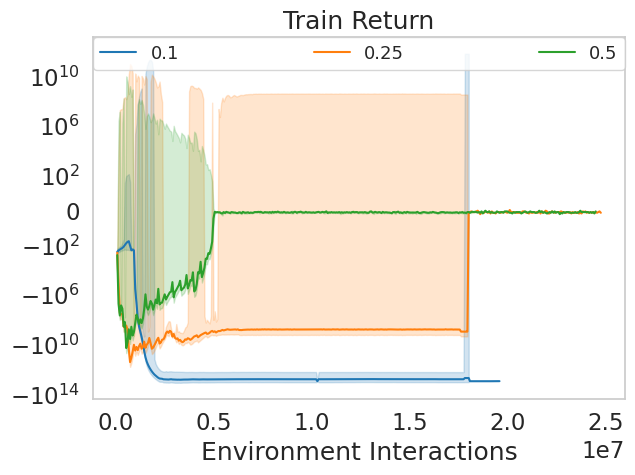}
    \subcaption{$N=100$}
\end{subfigure}
\caption{\textbf{Marginalized Gradient}: PPO training with marginalized gradients on HalfCheetah-v3. High variance gradient estimates lead to highly unstable learning dynamics.}
\label{fig:marginalized_gradient}
\end{center}
\vskip -0.2in
\end{figure}

We also experimented with a version of PPO that uses an estimate of the marginalized gradient in \cref{eq:marginalized gradient}.
As shown in \cref{fig:marginalized_gradient}, the large variance of the gradient estimate dominates the learning, even with 100 samples used for \cref{eq:marginalized gradient}.

\begin{figure*}[htb]
\vskip 0.2in
\begin{center}
\begin{subfigure}{.24\textwidth}
    \centering
    \includegraphics[width=\linewidth]{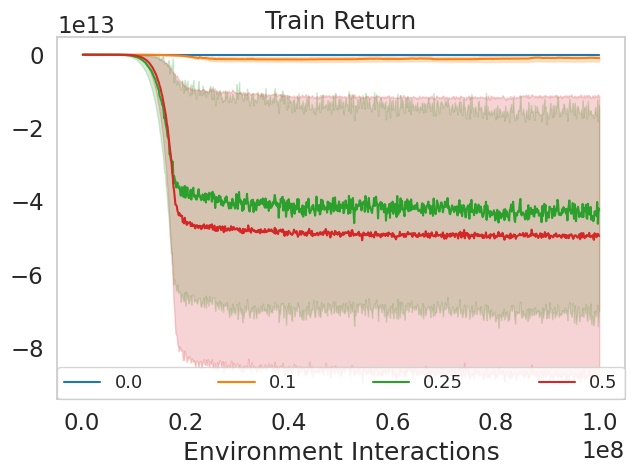}
    \subcaption{A2C}
\end{subfigure}
\begin{subfigure}{.24\textwidth}
    \centering
    \includegraphics[width=\linewidth]{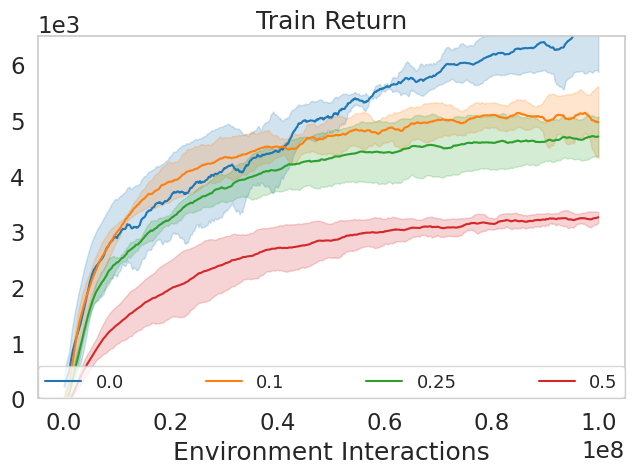}
    \subcaption{A2C Consistent}
\end{subfigure}
\begin{subfigure}{.24\textwidth}
    \centering
    \includegraphics[width=\linewidth]{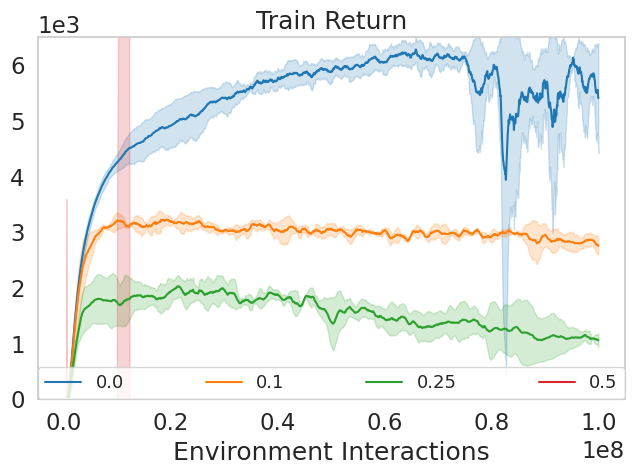}
    \subcaption{PPO}
\end{subfigure}
\begin{subfigure}{.24\textwidth}
    \centering
    \includegraphics[width=\linewidth]{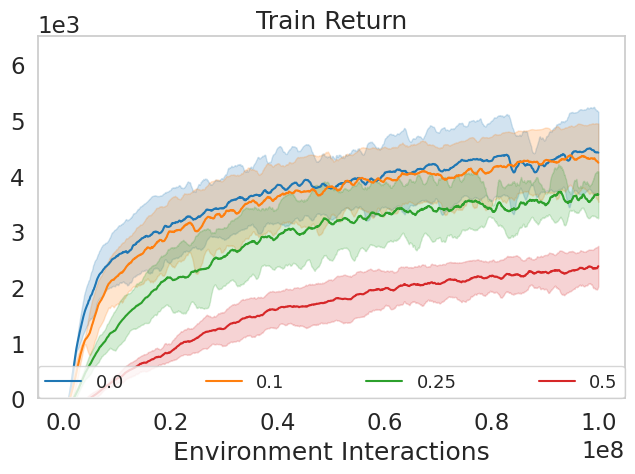}
    \subcaption{PPO Consistent}
\end{subfigure}
\caption{\textbf{MuJoCo-HalfCheetah-v3}: Clipped score of \textbf{A2C} and \textbf{PPO} with and without consistent dropout probabilities $0/0.1/0.25/0.5$. Curves are averages of three independent training runs.}
\label{fig:half_cheetah}
\end{center}
\vskip -0.2in
\end{figure*}

\begin{figure*}[htb]
\vskip 0.2in
\begin{center}
\begin{subfigure}{.24\textwidth}
    \centering
    \includegraphics[width=\linewidth]{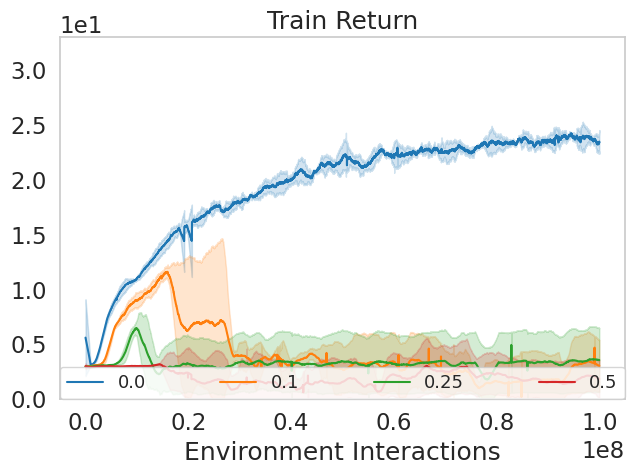}
    \subcaption{A2C}
\end{subfigure}
\begin{subfigure}{.24\textwidth}
    \centering
    \includegraphics[width=\linewidth]{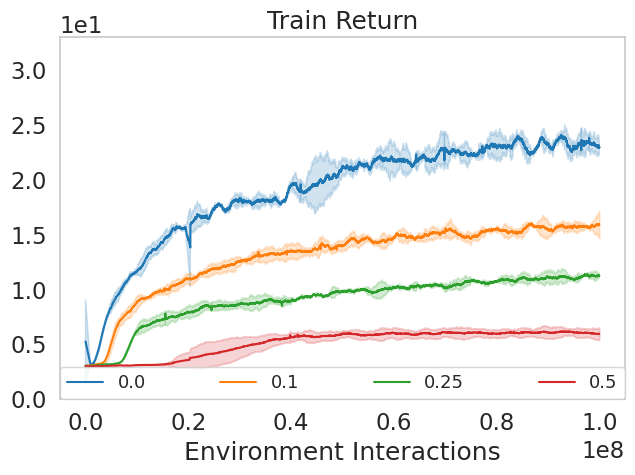}
    \subcaption{A2C Consistent}
\end{subfigure}
\begin{subfigure}{.24\textwidth}
    \centering
    \includegraphics[width=\linewidth]{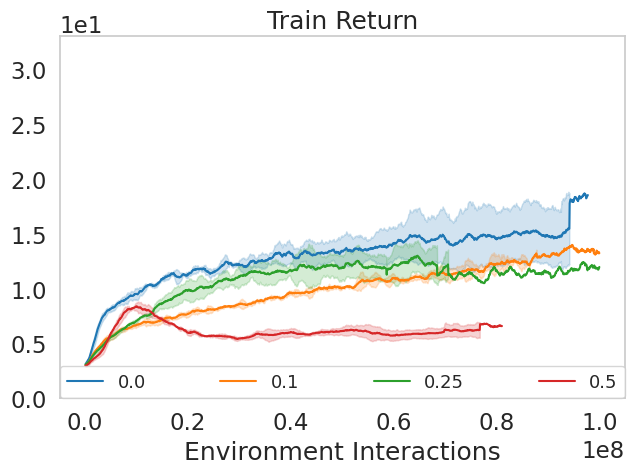}
    \subcaption{PPO}
\end{subfigure}
\begin{subfigure}{.24\textwidth}
    \centering
    \includegraphics[width=\linewidth]{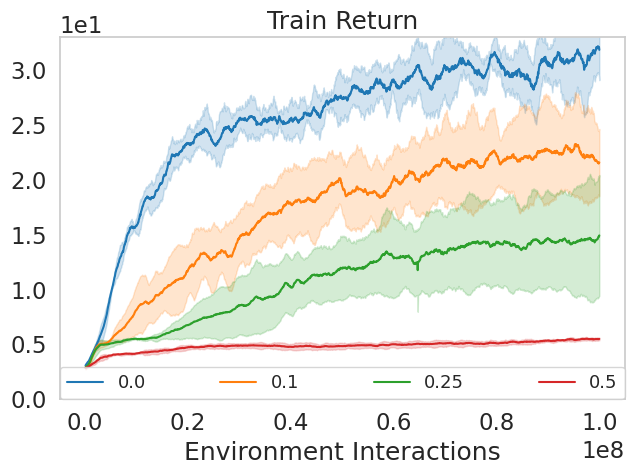}
    \subcaption{PPO Consistent}
\end{subfigure}
\caption{\textbf{Atari-Beamrider}: Clipped score of \textbf{A2C} and \textbf{PPO} with inconsistent and consistent dropout probabilities $0/0.1/0.25/0.5$. Curves reflect averages of three independent training runs.}
\label{fig:beamrider}
\end{center}
\vskip -0.2in
\end{figure*}

\subsection{Discrete-Action Environments}
\label{subsec:atari}

We tested three discrete-action Atari environments (Beamrider, Breakout, and Seaquest), and as shown in \cref{tab:results} A2C is intolerant to any amount of inconsistent dropout.
On the other hand, PPO often benefits from low-levels of inconsistent dropout (e.g., $0.1$). 
We believe the source of this benefit stems from an increase in policy entropy.
Nevertheless, inconsistent dropout triggers PPO's KL divergence-based early stopping, which prevents updates from occurring and precludes any policy learning.
Therefore, to facilitate training we disabled PPO's early stopping and forced a set number of gradient steps to be performed at each update.

Consistent dropout leads to stable learning (\cref{fig:beamrider}), but we observed that policies can become overly confident in their actions, as evidenced by quickly growing action probabilities and shrinking policy entropy.
Full Atari results may be found in Appendix \ref{sec:full_atari_results}.

\subsection{Stable GPT Training}

\begin{figure}[htb]
\begin{center}
\begin{subfigure}{.23\textwidth}
    \centering
    \includegraphics[width=\linewidth]{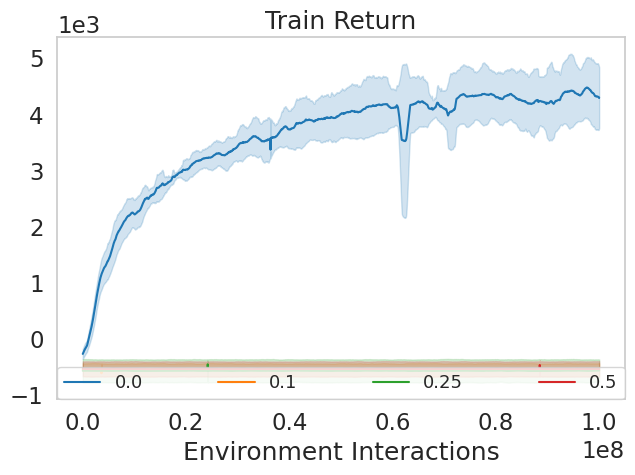}
    \subcaption{PPO}
\end{subfigure}
\begin{subfigure}{.23\textwidth}
    \centering
    \includegraphics[width=\linewidth]{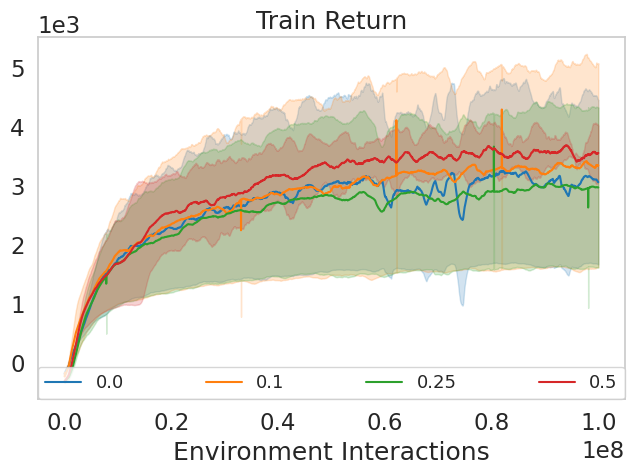}
    \subcaption{PPO Consistent}
\end{subfigure}
\caption{\textbf{GPT Training}: PPO training of GPT models on HalfCheetah-v3. With any level of inconsistent dropout all PPO models are unable to learn. Consistent dropout stabilizes training for all dropout probabilities.}
\label{fig:gpt_training}
\end{center}
\vskip -0.2in
\end{figure}

Prior work on training transformer models with reinforcement learning disable dropout \cite{parisotto19,loynd20}.
This is unsurprising as even small levels of dropout cause much larger changes to GPT's outputs compared to similarly deep MLP networks (Table \ref{tab:example}). 
As a result, PPO models trained with inconsistent dropout are unable to exceed random performance on Half-Cheetah.
On the other hand, consistent dropout (\cref{fig:gpt_training}) enables stable GPT training across all dropout probabilities.
Extrapolating from this result, we believe consistent dropout is applicable to a variety of network architectures.

\begin{figure}[htb]
\begin{center}
\begin{subfigure}{.23\textwidth}
    \centering
    \includegraphics[width=\linewidth]{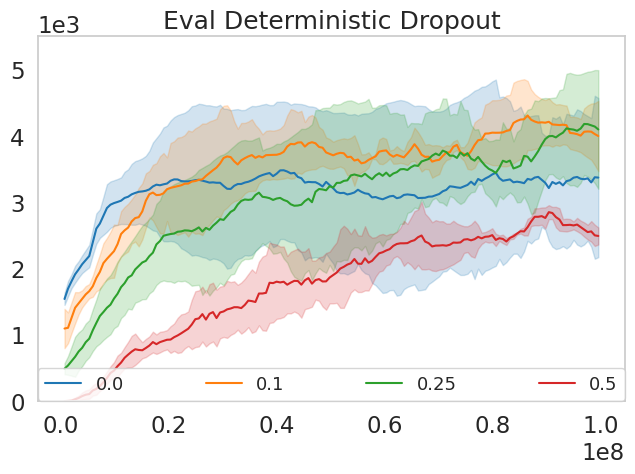}
    \subcaption{Dropout}
\end{subfigure}
\begin{subfigure}{.23\textwidth}
    \centering
    \includegraphics[width=\linewidth]{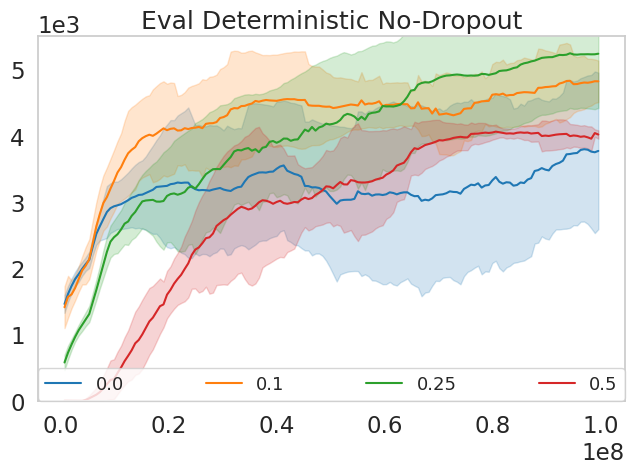}
    \subcaption{No Dropout}
\end{subfigure}
\caption{\textbf{Evaluating with Dropout}: Evaluation performance of Consistent PPO on Half-Cheetah-v3 with dropout enabled and disabled for dropout probabilities $0/0.1/0.25/0.5$. Similar to supervised learning, disabling dropout at evaluation time increases evaluation performance, particularly with a dropout of $0.5$. Curves reflect evaluations performed on models from three independent training runs.}
\label{fig:evaluation}
\end{center}
\vskip -0.2in
\end{figure}

\section{Evaluating with Dropout}
\label{sec:evaluation}

Common practice for supervised learning is to disable dropout at evaluation time.
However, it is unclear whether this practice would be beneficial for sequential decision making.
To answer this question, we examined the effects of enabling or disabling dropout at evaluation time, and found a strong preference towards disabling dropout at evaluation time (\cref{fig:evaluation}).
Specifically we observe a $\mathbf{2\%}$ improvement on average in final evaluation score when disabling dropout at evaluation time for the $p=0.1$ models, a $\mathbf{17\%}$ improvement for the $p=0.25$ models, and a $\mathbf{46\%}$ improvement in score for the $p=0.5$ models.
Finally, we find that models trained with inconsistent dropout are largely unaffected by enabling or disabling dropout at evaluation time.
Further evaluation results may be found in \cref{sec:full_mujoco_results}.


\section{Limitations}
\label{sec:limitations}

This study pertains only to on-policy policy gradient methods. Other types of reinforcement learning algorithms such as off-policy and model-based reinforcement learning are affected by dropout in different ways, which are beyond the scope of this paper.

Having stabilized dropout for policy gradient algorithms, our findings raise the larger question of whether dropout aids generalization to unseen tasks or novel variations of training environments.
Related work \citep{cobbe18, farebrother18} has found beneficial generalization effects from small amounts of dropout, and its common use in supervised settings suggests that dropout may aid generalization especially when training overparameterized networks.
However, we expect that thoroughly exploring how dropout contributes to generalization in sequential decision settings is a substantial undertaking, one that lies beyond the scope of this work.

\section{Conclusions}
\label{sec:conclusions}

We demonstrate that na\"ive application of dropout to policy gradient reinforcement learning results in instability due to the mismatch between the dropout masks that are applied during rollouts and those used during updates. 
We address this instability through consistent dropout, a simple approach for saving and reapplying the same dropout masks during rollouts and updates.
We demonstrate that this solution is both theoretically sound and empirically stable on a variety of continuous and discrete actions environments.
Additionally, we show that this technique scales to enable stable online training of transformer models such as GPT.
Given the complete lack of support for dropout in open-source reinforcement learning frameworks, we hope this work will pave the way for a re-evaluation of this time-tested technique.

\bibliography{references}
\bibliographystyle{icml2022}

\newpage
\appendix
\onecolumn

\section{Hyperparameters}
\label{sec:hyperparameters}
\begin{table*}[htb!]
\caption{\textbf{Hyperparameters:} Networks consisted of 3-layer MLPs with ReLU nonlinearities. Dropout was applied between each layer of the network. Note that PPO Consistent uses a target-KL early stopping condition to terminate updates when the KL divergence passes a threshold. We attempted to run standard PPO with the same parameter but found that it would trigger early termination due to the large KL divergences caused by inconsistent dropout masks. Thus, the best performance for standard PPO was with the target-KL disabled.}
\label{tab:hyperparams}
\vskip 0.15in
\begin{center}
\begin{scriptsize}
\begin{sc}
\begin{tabular}{crll}
\toprule
Variant & Hyperparameter & MuJoCo & Atari \\
\midrule
A2C & max steps & 1e8 & 1e8 \\
\& & workers & 16 & 16 \\
A2C Consistent & steps per epoch & 80 & 5 \\
 & learning rate & 7e-4 & 1e-4 \\
 & critic lr & 7e-4 & - \\
 & entropy coefficient & 0.01 & 0.01 \\
 & discount factor & 0.99 & 0.99 \\
 & gae-lambda & 0.95 & 0.95 \\
 & hidden size & 64 & 512 \\
 & gradient norm clip & 0.5 & 0.5 \\
 & rms prop epsilon & 3e-6 & 3e-6 \\
 & advantage normalization & 1 & 0 \\
\hline
PPO & max steps & 1e8 & 1e8 \\
 & workers & 16 & 16 \\
 & learning rate & 3e-4 & 1e-4 \\
 & steps per epoch & 4096 & 4096 \\
 & gradient steps per update & 16 & 16 \\
 & batch size & 64 & 64 \\
 & hidden size & 64 & 512 \\
 & value coefficient & 0.5 & 0.5 \\
 & entropy coefficient & 0.01 & 0 \\
 & gradient norm clip & 0.5 & 0.5 \\
 & discount factor & 0.99 & 0.99 \\
 & gae-lambda & 0.97 & 0.95 \\
 & clip ratio & 0.2 & 0.2 \\
 & target KL & - & - \\
\hline
PPO Consistent & max steps & 1e8 & 1e8 \\
 & workers & 16 & 16 \\
 & learning rate & 3e-4 & 1e-4 \\
 & steps per epoch & 4096 & 4096\\
 & gradient steps per update & 16 & 16\\
 & batch size & 64 & 64 \\
 & hidden size & 64 & 512 \\
 & value coefficient & 0.5 & 0.5 \\
 & entropy coefficient & 0.01 & 0 \\
 & gradient norm clip & 0.5 & 0.5 \\
 & discount factor & 0.99 & 0.99 \\
 & gae-lambda & 0.97 & 0.95 \\
 & clip ratio & 0.2 & 0.2 \\
 & target KL & 0.01 & 0.01 \\
\hline
GPT & max steps & 1e8 & \\
\& & workers & 16 & \\
GPT Consistent & learning rate & 3e-4 & \\
 & critic learning rate & 7e-4 & \\
 & steps per epoch & 1024 & \\
 & gradient steps per update & 128 \\
 & batch size & 64 & \\
 & hidden size & 64 & \\
 & entropy coefficient & 0.01 & \\
 & gradient norm clip & 0.5 & \\
 & discount factor & 0.99 & \\
 & gae-lambda & 0.97 & \\
 & clip ratio & 0.2 & \\
 & target KL & 0.01 & \\
 & block size & 8 & \\
 & number of layers & 4 & \\
 & number of heads & 4 & \\
\bottomrule
\end{tabular}
\end{sc}
\end{scriptsize}
\end{center}
\vskip -0.1in
\end{table*}

\newpage
\section{RL Framework Testing}
\label{sec:framework_testing}

The following tables show results from testing \textit{inconsistent dropout} on three open-source reinforcement learning frameworks. This was accomplished by adding dropout layers to the policy networks and running the A2C and PPO on the HalfCheetah-v3 environment using the hyperparameters suggested by the respective codebase.
In all the following tables, where ReLU was not used, Tanh was applied.

\begin{table*}[htb!]
\caption{Tianshou (version 0.4.2) on HalfCheetah-v3: For A2C, ReLU activation seems to be the biggest causative factor of instability, while Tanh is a stabilizer. ActionClipping can also help stabilize when dropout levels are moderate. Reward normalization may have a slightly helpful effect, but less than other factors. PPO even with moderate levels of dropout demonstrated broad instability: PPO$^\dagger$ indicates early termination of the experiment due to instabilities resulting in NaN values.}
\label{tab:tianshou_results}
\vskip 0.15in
\begin{center}
\begin{small}
\begin{sc}
\begin{tabular}{lcccccc}
\toprule
Alg & Dropout & RewNorm? & ActClip? & ReLU? & Best & Final \\
\midrule
A2C & $0$ & \cmark & \cmark & \xmark & $1556$ & $1542$ \\
A2C & $0$ & \cmark & \cmark & \cmark & $2949$ & $2927$ \\
A2C & $0.1$ & \xmark & \cmark & \xmark & $3192$ & $2969$ \\
A2C & $0.1$ & \xmark & \cmark & \cmark & $3085$ & $2693$ \\
A2C & $0.1$ & \xmark & \xmark & \xmark & $3593$ & $3357$ \\
A2C & $0.1$ & \xmark & \xmark & \cmark & $1628$ & $-3\mathrm{e}7$ \\
A2C & $0.1$ & \cmark & \cmark & \xmark & $1193$ & $1152$ \\
A2C & $0.1$ & \cmark & \cmark & \cmark & $537$ & $573$ \\
A2C & $0.1$ & \cmark & \xmark & \xmark & $3066$ & $2900$ \\
A2C & $0.1$ & \cmark & \xmark & \cmark & $3204$ & $-1.2\mathrm{e}8$ \\
A2C & $0.25$ & \xmark & \cmark & \xmark & $1049$ & $977$ \\
A2C & $0.25$ & \xmark & \cmark & \cmark & $171$ & $-13$ \\
A2C & $0.25$ & \xmark & \xmark & \xmark & $1065$ & $-3.4\mathrm{e}7$ \\
A2C & $0.25$ & \xmark & \xmark & \cmark & $-146$ & $-3.9\mathrm{e}9$ \\
A2C & $0.25$ & \cmark & \cmark & \xmark & $1141$ & $1139$ \\
A2C & $0.25$ & \cmark & \cmark & \cmark & $1137$ & $1048$ \\
A2C & $0.25$ & \cmark & \xmark & \xmark & $3872$ & $3313$ \\
A2C & $0.25$ & \cmark & \xmark & \cmark & $260$ & $-2.1\mathrm{e}9$ \\
\hline
PPO & $0$ & \cmark & \cmark & \xmark & $7472$ & $5645$ \\
PPO & $0$ & \cmark & \cmark & \cmark & $8743$ & $7890$ \\
PPO$^\dagger$ & $0.1$ & \xmark & \cmark & \xmark & $1708$ & $-335$ \\
PPO$^\dagger$ & $0.1$ & \xmark & \cmark & \cmark & $1548$ & $463$ \\
PPO$^\dagger$ & $0.1$ & \xmark & \xmark & \xmark & $2267$ & $1162$ \\
PPO$^\dagger$ & $0.1$ & \xmark & \xmark & \cmark & $1372$ & $1050$ \\
PPO$^\dagger$ & $0.1$ & \cmark & \cmark & \xmark & $2238$ & $1813$ \\
PPO$^\dagger$ & $0.1$ & \cmark & \cmark & \cmark & $408$ & $42$ \\
PPO$^\dagger$ & $0.1$ & \cmark & \xmark & \xmark & $2268$ & $1870$ \\
PPO$^\dagger$ & $0.1$ & \cmark & \xmark & \cmark & $1007$ & $-55$ \\
PPO$^\dagger$ & $0.25$ & \xmark & \cmark & \xmark & $1001$ & $881$ \\
PPO$^\dagger$ & $0.25$ & \xmark & \cmark & \cmark & $279$ & $153$ \\
PPO$^\dagger$ & $0.25$ & \xmark & \xmark & \xmark & $1270$ & $1006$ \\
PPO$^\dagger$ & $0.25$ & \xmark & \xmark & \cmark & $614$ & $246$ \\
PPO$^\dagger$ & $0.25$ & \cmark & \cmark & \xmark & $1292$ & $1167$ \\
PPO$^\dagger$ & $0.25$ & \cmark & \cmark & \cmark & $705$ & $-123$ \\
PPO$^\dagger$ & $0.25$ & \cmark & \xmark & \xmark & $348$ & $1321$ \\
PPO$^\dagger$ & $0.25$ & \cmark & \xmark & \cmark & $348$ & $-160$ \\
\bottomrule
\end{tabular}
\end{sc}
\end{small}
\end{center}
\vskip -0.1in
\end{table*}


\begin{table*}[htb!]
\caption{Stable-Baselines3 (version 1.2.0a0) on HalfCheetah-v3. Policy loss is shown in to provide an intuition on about the stability of the training process. Very high policy losses correlate to high-magnitude $\log\pi(a|s)$ values, a symptom of unstable dynamics induced by inconsistent dropout.}
\label{tab:sb3_results}
\vskip 0.15in
\begin{center}
\begin{small}
\begin{sc}
\begin{tabular}{lccccc}
\toprule
Alg & Dropout & ReLU? & Ortho Init? & Final & Policy Loss \\
\midrule
A2C & $0$ & \xmark & \xmark & $974$ & $2.5$ \\
A2C & $0$ & \xmark & \cmark & $688$ & $-1.1$ \\
A2C & $0$ & \cmark & \xmark & $526$ & $8$ \\
A2C & $0$ & \cmark & \cmark & $1954$ & $-5.4$ \\
A2C & $0.1$ & \xmark & \xmark & $1003$ & $-12.1$ \\
A2C & $0.1$ & \xmark & \cmark & $650$ & $-8.8$ \\
A2C & $0.1$ & \cmark & \xmark & $1639$ & $8.8\mathrm{e}3$ \\
A2C & $0.1$ & \cmark & \cmark & $-40$ & $-242$ \\
A2C & $0.25$ & \xmark & \xmark & $1325$ & $26$ \\
A2C & $0.25$ & \xmark & \cmark & $499$ & $25$ \\
A2C & $0.25$ & \cmark & \xmark & $1497$ & $-2.9\mathrm{e}5$ \\
A2C & $0.25$ & \cmark & \cmark & $950$ & $-2.7\mathrm{e}5$ \\
A2C & $0.5$ & \xmark & \xmark & $116$ & $1\mathrm{e}3$ \\
A2C & $0.5$ & \xmark & \cmark & $639$ & $-7.2\mathrm{e}3$ \\
A2C & $0.5$ & \cmark & \xmark & $1527$ & $1.7\mathrm{e}8$ \\
A2C & $0.5$ & \cmark & \cmark & $1083$ & $-7.8\mathrm{e}8$ \\
\hline
PPO & $0$ & \xmark & \xmark & $2238$ & $-0.007$ \\
PPO & $0$ & \xmark & \cmark & $3211$ & $-0.001$ \\
PPO & $0$ & \cmark & \xmark & $1232$ & $-0.0006$ \\
PPO & $0$ & \cmark & \cmark & $1277$ & $-0.001$ \\
PPO & $0.1$ & \xmark & \xmark & $1963$ & $0.04$ \\
PPO & $0.1$ & \xmark & \cmark & $1256$ & $0.05$ \\
PPO & $0.1$ & \cmark & \xmark & $1830$ & $0.04$ \\
PPO & $0.1$ & \cmark & \cmark & $2463$ & $0.04$ \\
PPO & $0.25$ & \xmark & \xmark & $1468$ & $0.04$ \\
PPO & $0.25$ & \xmark & \cmark & $1383$ & $0.04$ \\
PPO & $0.25$ & \cmark & \xmark & $2217$ & $0.05$ \\
PPO & $0.25$ & \cmark & \cmark & $1825$ & $0.05$ \\
PPO & $0.5$ & \xmark & \xmark & $1331$ & $0.06$ \\
PPO & $0.5$ & \xmark & \cmark & $1328$ & $0.06$ \\
PPO & $0.5$ & \cmark & \xmark & $904$ & $0.06$ \\
PPO & $0.5$ & \cmark & \cmark & $993$ & $0.05$ \\
\bottomrule
\end{tabular}
\end{sc}
\end{small}
\end{center}
\vskip -0.1in
\end{table*}

\begin{table*}[htb!]
\caption{RLlib (version 1.8.0) on HalfCheetah-v3. Dagger denotes training became numerically unstable and produced errors. For these runs, we report the final performance level before the error.}
\label{tab:rllib_results}
\vskip 0.15in
\begin{center}
\begin{small}
\begin{sc}
\begin{tabular}{lccc}
\toprule
Alg & Dropout & ReLU? & Final \\
\midrule
A2C & $0$ & \xmark & $712$ \\
A2C & $0$ & \cmark & $747$ \\
A2C & $0.1$ & \xmark & $-393$ \\
A2C$^\dagger$ & $0.1$ & \cmark & $-439$ \\
A2C$^\dagger$ & $0.25$ & \xmark & $-253$ \\
A2C$^\dagger$ & $0.25$ & \cmark & $-594$ \\
A2C$^\dagger$ & $0.5$ & \xmark & $-643$ \\
A2C$^\dagger$ & $0.5$ & \cmark & $-596$ \\
\hline
PPO & $0$ & \xmark & $2044$ \\
PPO & $0$ & \cmark & $5390$ \\
PPO & $0.1$ & \xmark & $909$ \\
PPO$^\dagger$ & $0.1$ & \cmark & $-109$ \\
PPO & $0.25$ & \xmark & $383$ \\
PPO$^\dagger$ & $0.25$ & \cmark & $-238$ \\
PPO & $0.5$ & \xmark & $867$ \\
PPO & $0.5$ & \cmark & $1517$ \\
\bottomrule
\end{tabular}
\end{sc}
\end{small}
\end{center}
\vskip -0.1in
\end{table*}

\clearpage
\section{Full MuJoCo Results}
\label{sec:full_mujoco_results}

Hyperparameters were tuned on Half-Cheetah-v3 then deployed without further modification on Hopper-v3 and Walker2d-v3.

\begin{figure*}[htb!]
\begin{center}
\begin{subfigure}{.24\textwidth}
    \centering
    \includegraphics[width=\linewidth]{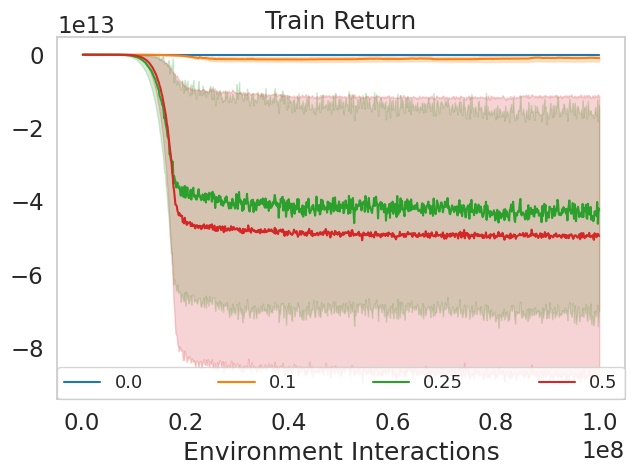}
    \subcaption[]{A2C}
\end{subfigure}
\begin{subfigure}{.24\textwidth}
    \centering
    \includegraphics[width=\linewidth]{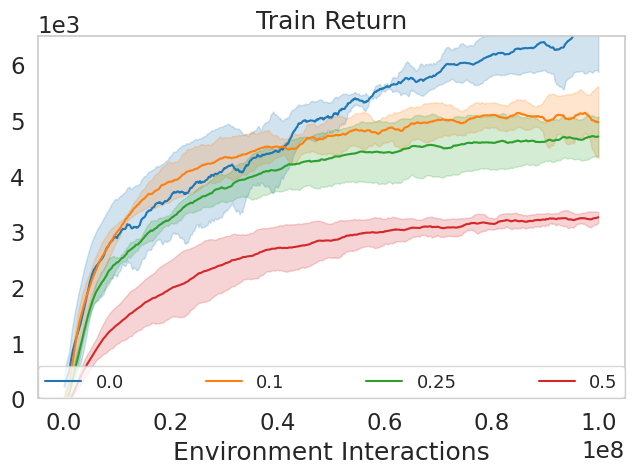}
    \subcaption[]{A2C Consistent}
\end{subfigure}
\begin{subfigure}{.24\textwidth}
    \centering
    \includegraphics[width=\linewidth]{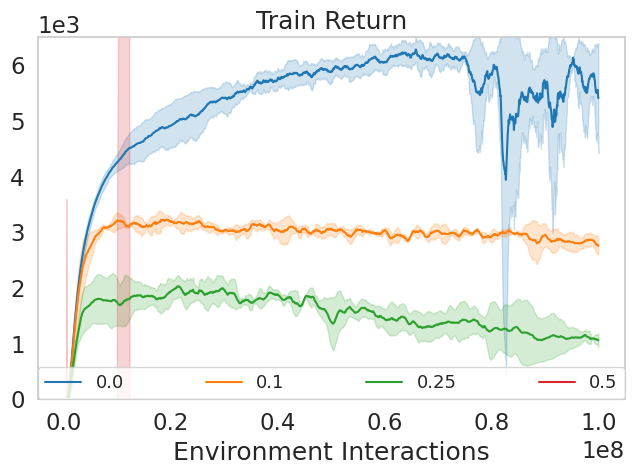}
    \subcaption[]{PPO}
\end{subfigure}
\begin{subfigure}{.24\textwidth}
    \centering
    \includegraphics[width=\linewidth]{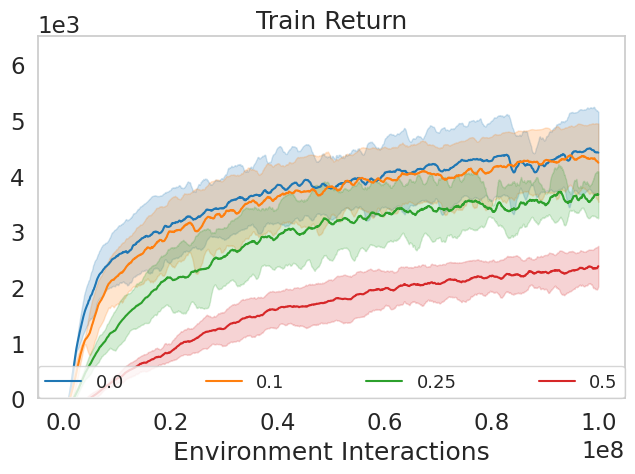}
    \subcaption[]{PPO Consistent}
\end{subfigure}
\caption{\textbf{Half-Cheetah-v3}: Training performance of various models on MuJoCo Half-Cheetah-v3. Curves are averages of three independent training runs.}
\label{fig:half_cheetah_train}
\end{center}
\vskip -0.2in
\end{figure*}

\begin{figure*}[htb!]
\begin{center}
\begin{subfigure}{.24\textwidth}
    \centering
    \includegraphics[width=\linewidth]{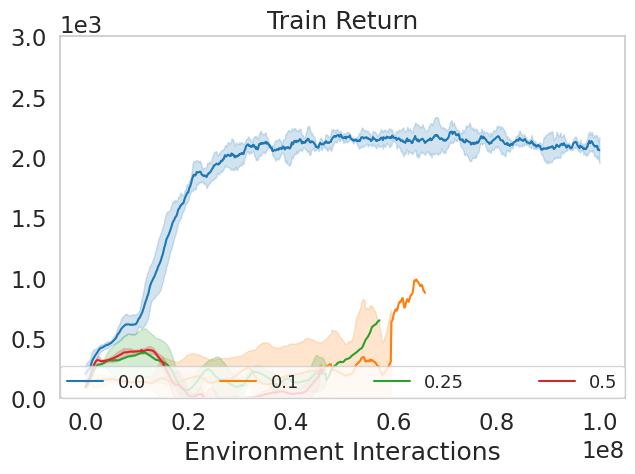}
    \subcaption[]{A2C}
\end{subfigure}
\begin{subfigure}{.24\textwidth}
    \centering
    \includegraphics[width=\linewidth]{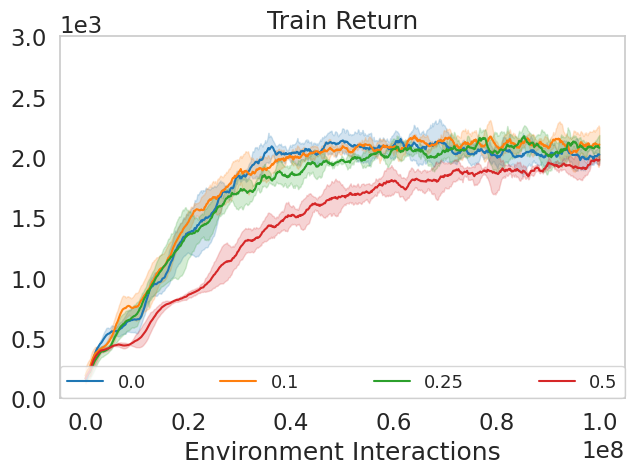}
    \subcaption[]{A2C Consistent}
\end{subfigure}
\begin{subfigure}{.24\textwidth}
    \centering
    \includegraphics[width=\linewidth]{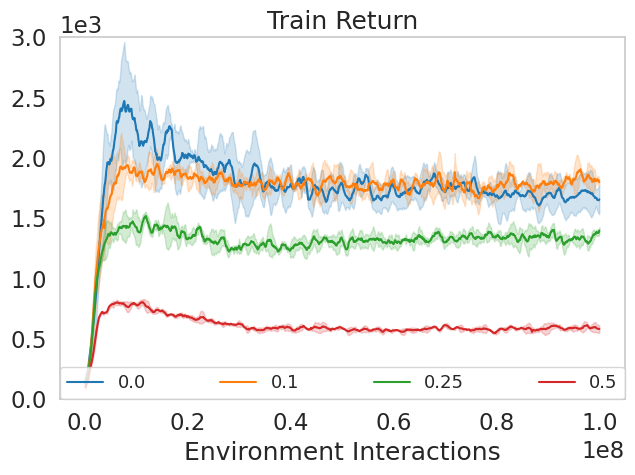}
    \subcaption[]{PPO}
\end{subfigure}
\begin{subfigure}{.24\textwidth}
    \centering
    \includegraphics[width=\linewidth]{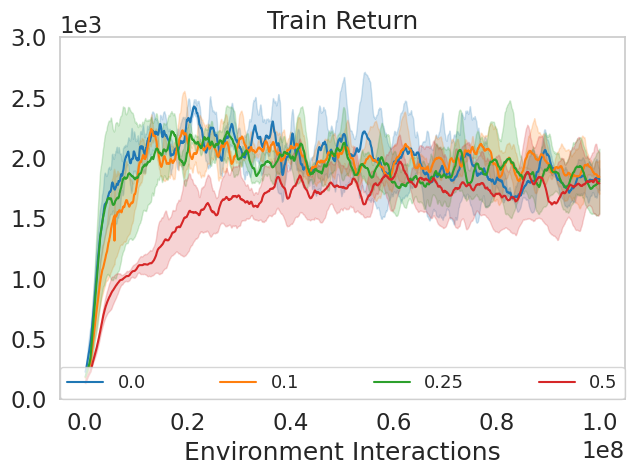}
    \subcaption[]{PPO Consistent}
\end{subfigure}
\caption{\textbf{Hopper-v3}: Training performance of various models on MuJoCo Hopper-v3. Curves are averages of three independent training runs.}
\label{fig:hopper_train}
\end{center}
\vskip -0.2in
\end{figure*}

\begin{figure*}[htb!]
\begin{center}
\begin{subfigure}{.24\textwidth}
    \centering
    \includegraphics[width=\linewidth]{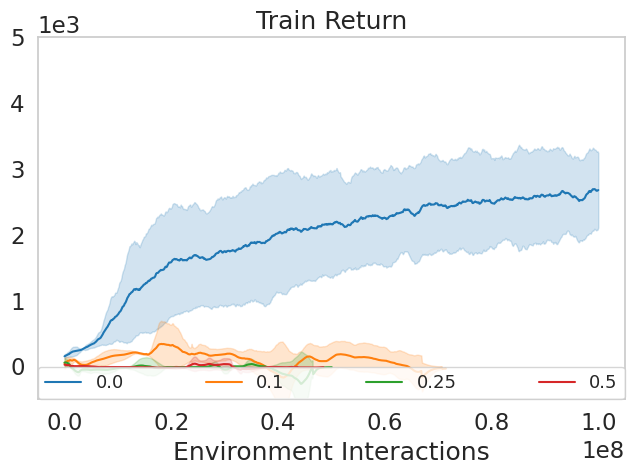}
    \subcaption[]{A2C}
\end{subfigure}
\begin{subfigure}{.24\textwidth}
    \centering
    \includegraphics[width=\linewidth]{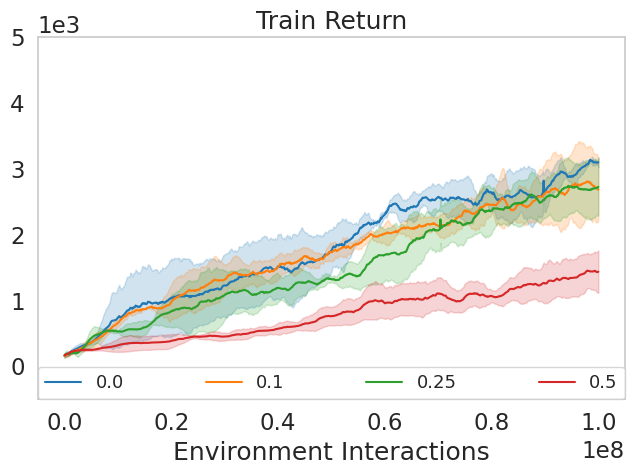}
    \subcaption[]{A2C Consistent}
\end{subfigure}
\begin{subfigure}{.24\textwidth}
    \centering
    \includegraphics[width=\linewidth]{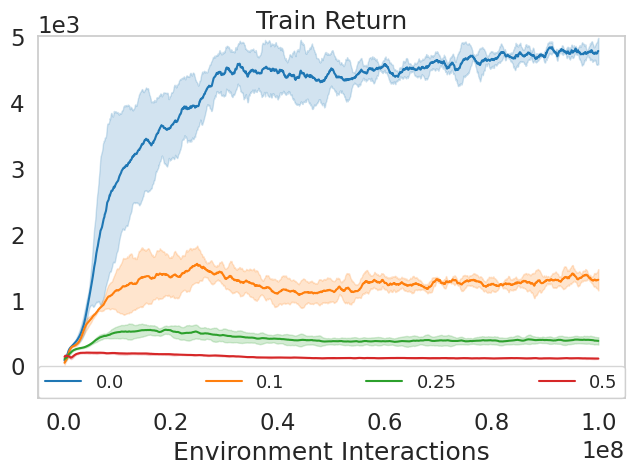}
    \subcaption[]{PPO}
\end{subfigure}
\begin{subfigure}{.24\textwidth}
    \centering
    \includegraphics[width=\linewidth]{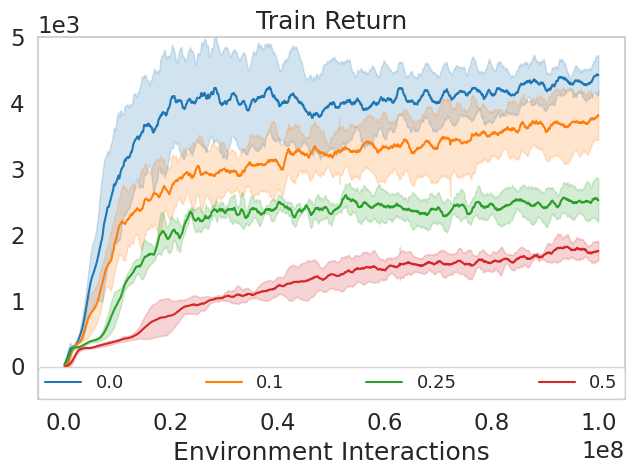}
    \subcaption[]{PPO Consistent}
\end{subfigure}
\caption{\textbf{Walker2d-v3}: Training performance of various models on MuJoCo Walker2d-v3. Curves are averages of three independent training runs.}
\label{fig:walker2d_train}
\end{center}
\vskip -0.2in
\end{figure*}

\begin{figure*}[htb!]
\begin{center}
\begin{subfigure}{.24\textwidth}
    \centering
    \includegraphics[width=\linewidth]{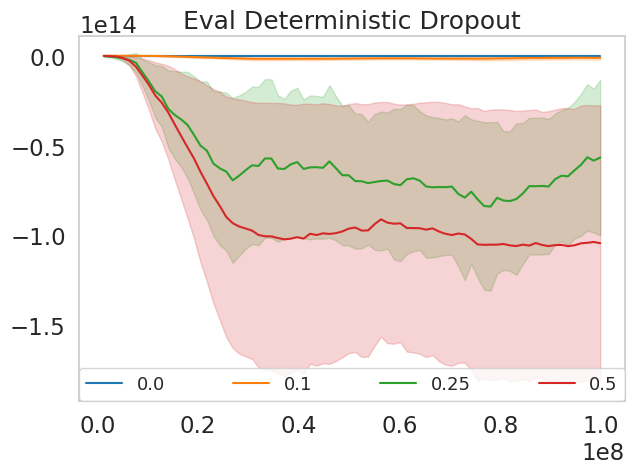}
    \subcaption[]{A2C}
\end{subfigure}
\begin{subfigure}{.24\textwidth}
    \centering
    \includegraphics[width=\linewidth]{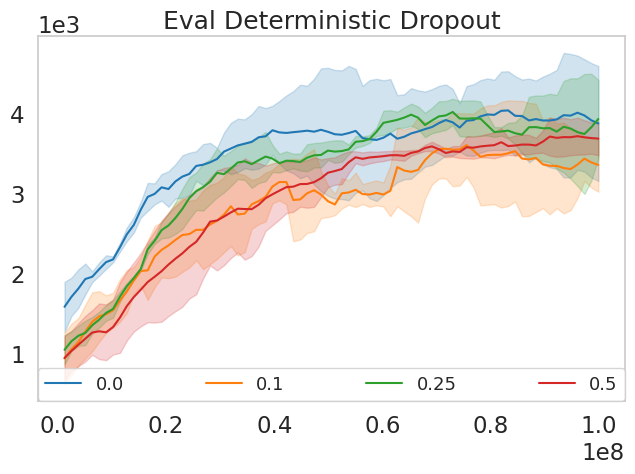}
    \subcaption[]{A2C Consistent}
\end{subfigure}
\begin{subfigure}{.24\textwidth}
    \centering
    \includegraphics[width=\linewidth]{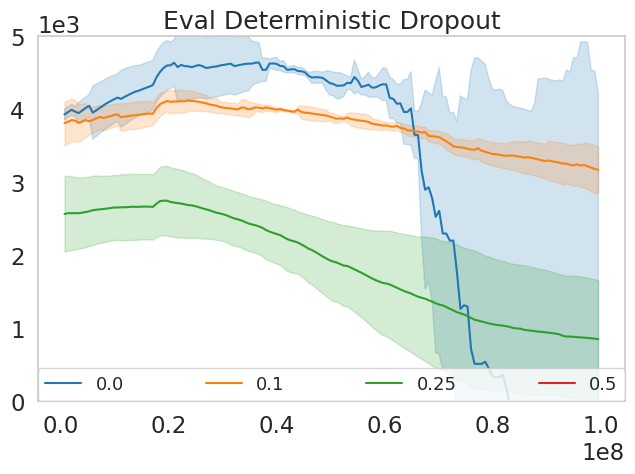}
    \subcaption[]{PPO}
\end{subfigure}
\begin{subfigure}{.24\textwidth}
    \centering
    \includegraphics[width=\linewidth]{figures/mujoco/half_cheetah/ppo_s_eval_det_drop.png}
    \subcaption[]{PPO Consistent}
\end{subfigure}
\caption{\textbf{Half-Cheetah-v3 Deterministic Evaluation with Dropout}: Curves are averages of three independent training runs.}
\label{fig:half_cheetah_eval_det_drop}
\end{center}
\vskip -0.2in
\end{figure*}

\begin{figure*}[htb!]
\begin{center}
\begin{subfigure}{.24\textwidth}
    \centering
    \includegraphics[width=\linewidth]{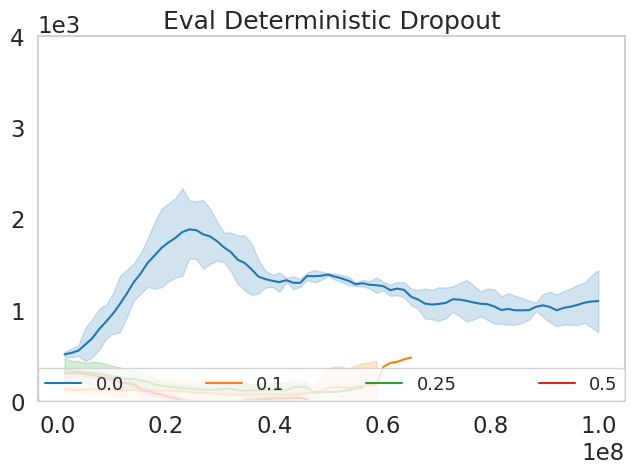}
    \subcaption[]{A2C}
\end{subfigure}
\begin{subfigure}{.24\textwidth}
    \centering
    \includegraphics[width=\linewidth]{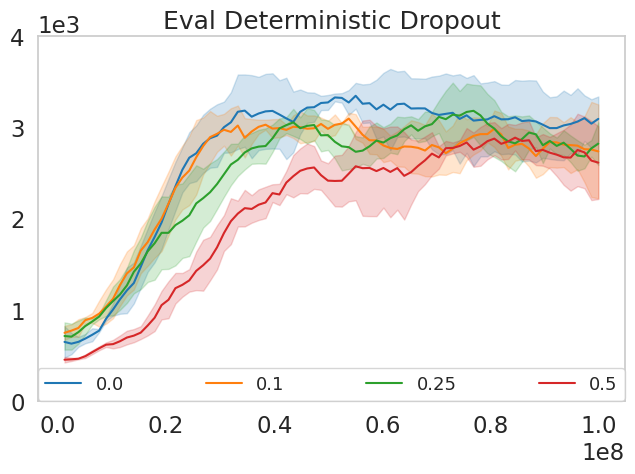}
    \subcaption[]{A2C Consistent}
\end{subfigure}
\begin{subfigure}{.24\textwidth}
    \centering
    \includegraphics[width=\linewidth]{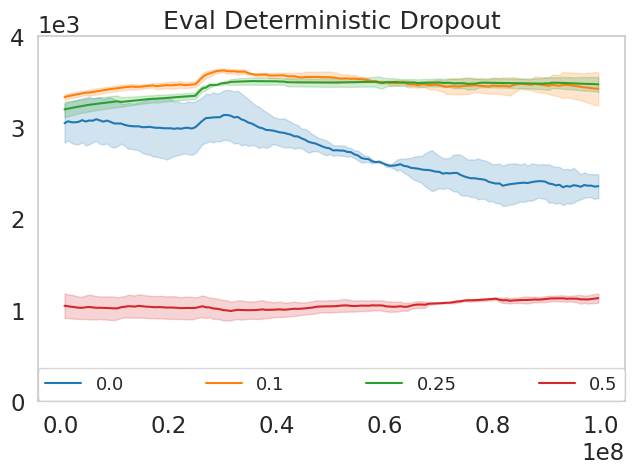}
    \subcaption[]{PPO}
\end{subfigure}
\begin{subfigure}{.24\textwidth}
    \centering
    \includegraphics[width=\linewidth]{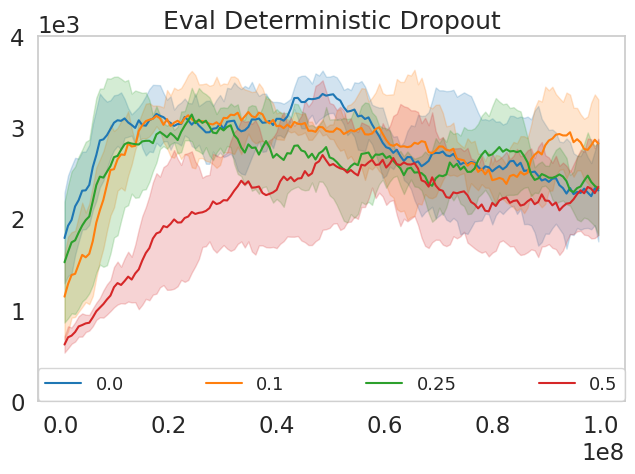}
    \subcaption[]{PPO Consistent}
\end{subfigure}
\caption{\textbf{Hopper-v3 Deterministic Evaluation with Dropout}: Curves are averages of three independent training runs.}
\label{fig:hopper_eval_det_drop}
\end{center}
\vskip -0.2in
\end{figure*}

\begin{figure*}[htb!]
\begin{center}
\begin{subfigure}{.24\textwidth}
    \centering
    \includegraphics[width=\linewidth]{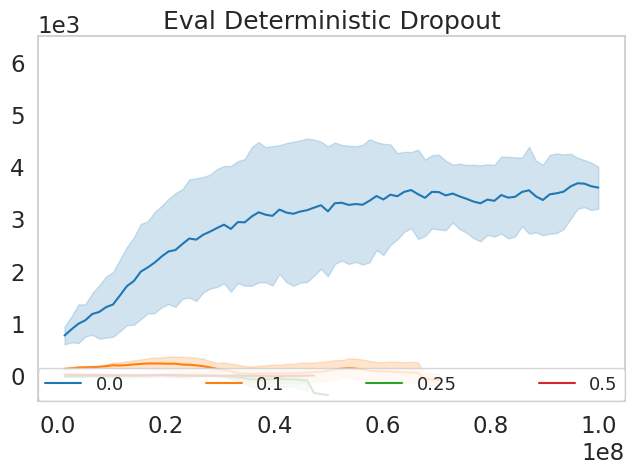}
    \subcaption[]{A2C}
\end{subfigure}
\begin{subfigure}{.24\textwidth}
    \centering
    \includegraphics[width=\linewidth]{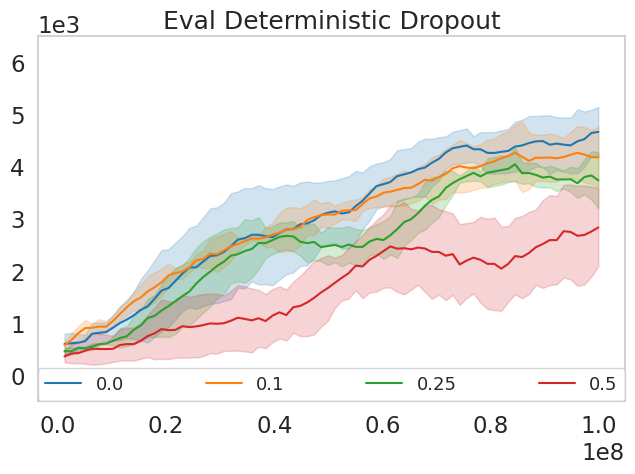}
    \subcaption[]{A2C Consistent}
\end{subfigure}
\begin{subfigure}{.24\textwidth}
    \centering
    \includegraphics[width=\linewidth]{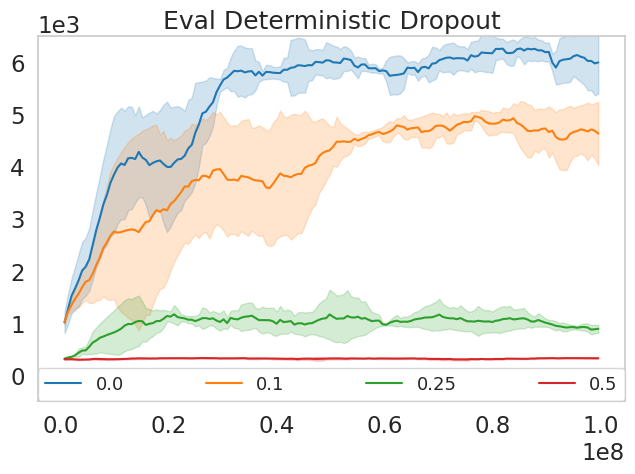}
    \subcaption[]{PPO}
\end{subfigure}
\begin{subfigure}{.24\textwidth}
    \centering
    \includegraphics[width=\linewidth]{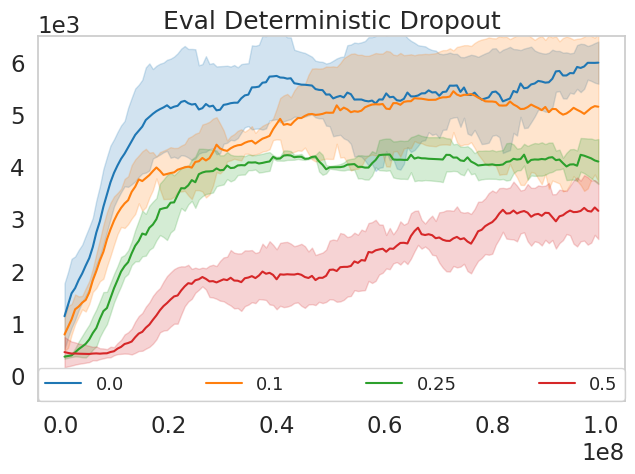}
    \subcaption[]{PPO Consistent}
\end{subfigure}
\caption{\textbf{Walker2d-v3 Deterministic Evaluation with Dropout}: Curves are averages of three independent training runs.}
\label{fig:walker2d_eval_det_drop}
\end{center}
\vskip -0.2in
\end{figure*}

\begin{figure*}[htb!]
\begin{center}
\begin{subfigure}{.24\textwidth}
    \centering
    \includegraphics[width=\linewidth]{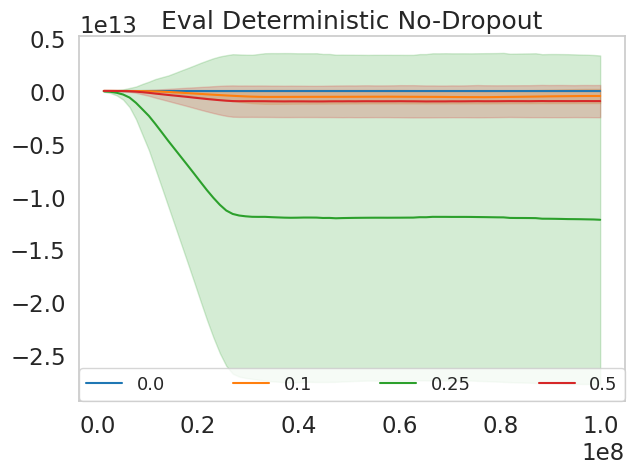}
    \subcaption[]{A2C}
\end{subfigure}
\begin{subfigure}{.24\textwidth}
    \centering
    \includegraphics[width=\linewidth]{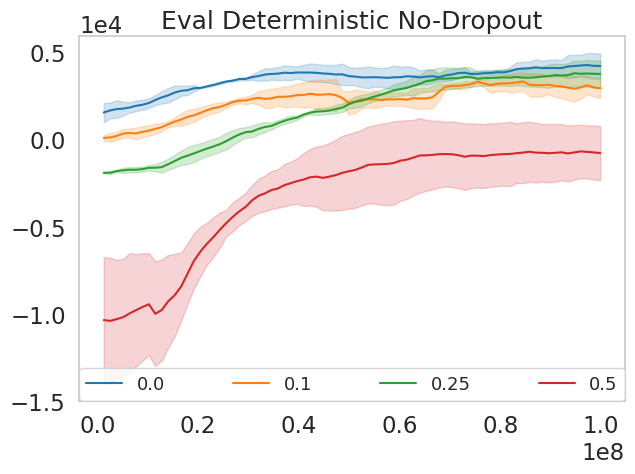}
    \subcaption[]{A2C Consistent}
\end{subfigure}
\begin{subfigure}{.24\textwidth}
    \centering
    \includegraphics[width=\linewidth]{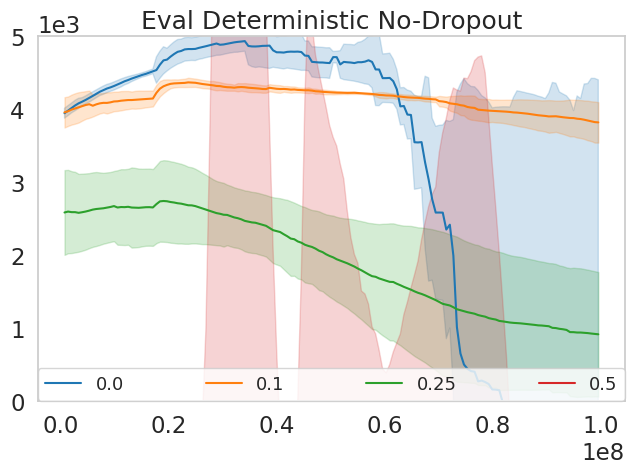}
    \subcaption[]{PPO}
\end{subfigure}
\begin{subfigure}{.24\textwidth}
    \centering
    \includegraphics[width=\linewidth]{figures/mujoco/half_cheetah/ppo_s_eval_det_nodrop.png}
    \subcaption[]{PPO Consistent}
\end{subfigure}
\caption{\textbf{Half-Cheetah-v3 Deterministic Evaluation No-Dropout}: Curves are averages of three independent training runs.}
\label{fig:half_cheetah_eval_det_nodrop}
\end{center}
\vskip -0.2in
\end{figure*}

\begin{figure*}[htb!]
\begin{center}
\begin{subfigure}{.24\textwidth}
    \centering
    \includegraphics[width=\linewidth]{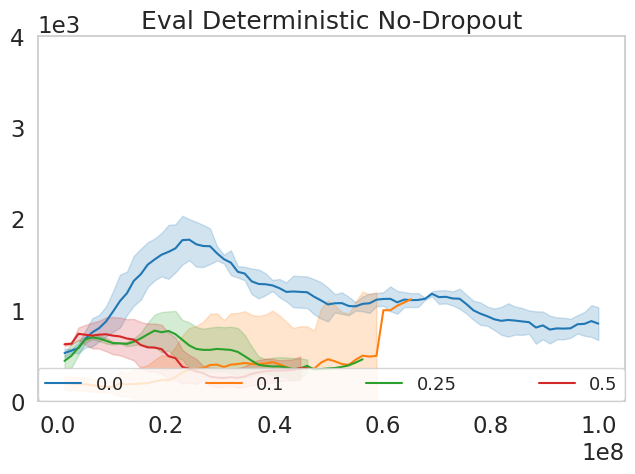}
    \subcaption[]{A2C}
\end{subfigure}
\begin{subfigure}{.24\textwidth}
    \centering
    \includegraphics[width=\linewidth]{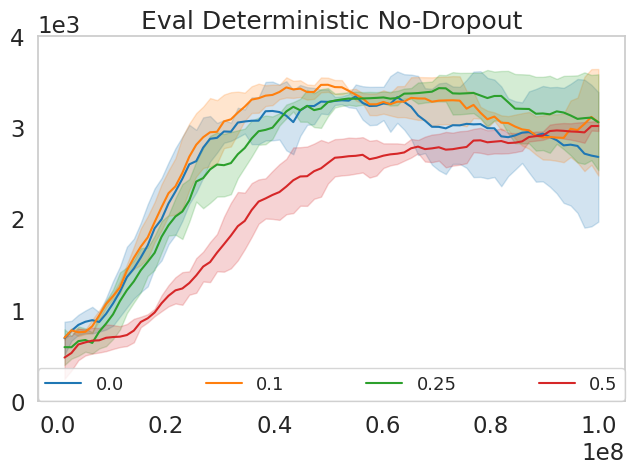}
    \subcaption[]{A2C Consistent}
\end{subfigure}
\begin{subfigure}{.24\textwidth}
    \centering
    \includegraphics[width=\linewidth]{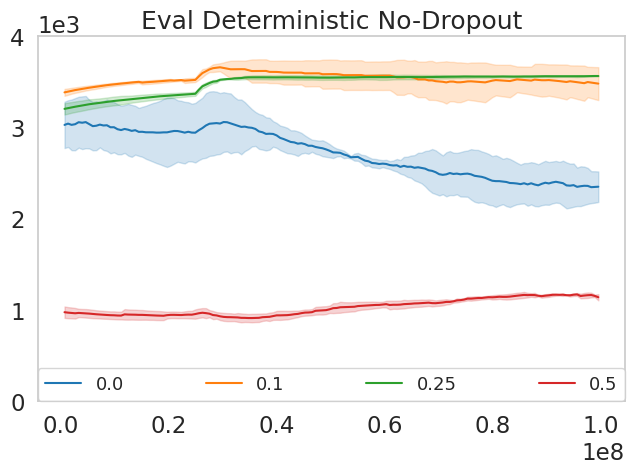}
    \subcaption[]{PPO}
\end{subfigure}
\begin{subfigure}{.24\textwidth}
    \centering
    \includegraphics[width=\linewidth]{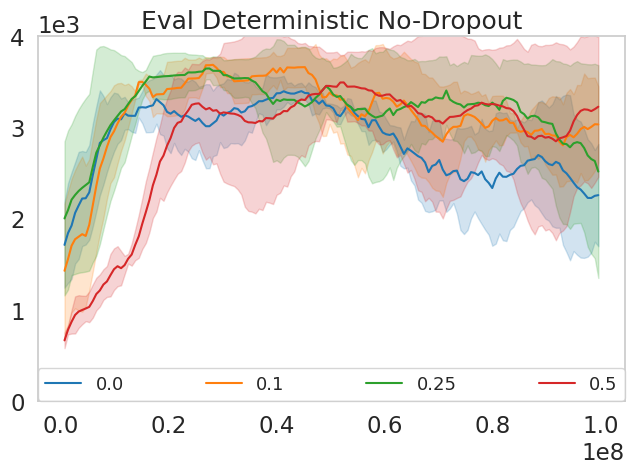}
    \subcaption[]{PPO Consistent}
\end{subfigure}
\caption{\textbf{Hopper-v3 Deterministic Evaluation No-Dropout}: Curves are averages of three independent training runs.}
\label{fig:hopper_eval_det_nodrop}
\end{center}
\vskip -0.2in
\end{figure*}

\begin{figure*}[htb!]
\begin{center}
\begin{subfigure}{.24\textwidth}
    \centering
    \includegraphics[width=\linewidth]{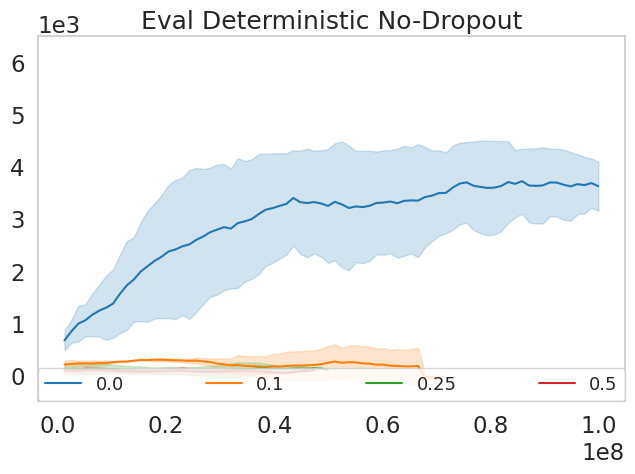}
    \subcaption[]{A2C}
\end{subfigure}
\begin{subfigure}{.24\textwidth}
    \centering
    \includegraphics[width=\linewidth]{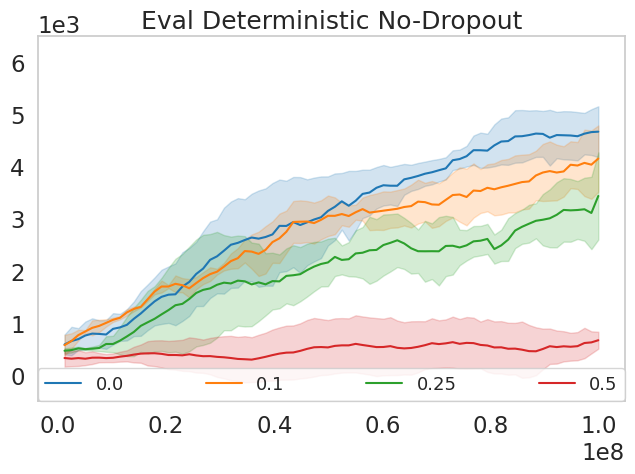}
    \subcaption[]{A2C Consistent}
\end{subfigure}
\begin{subfigure}{.24\textwidth}
    \centering
    \includegraphics[width=\linewidth]{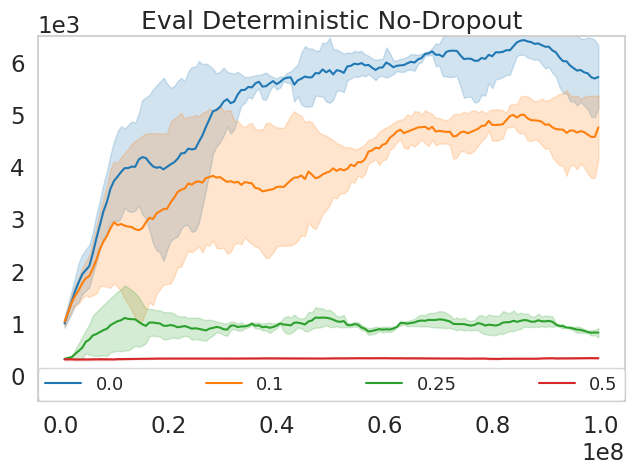}
    \subcaption[]{PPO}
\end{subfigure}
\begin{subfigure}{.24\textwidth}
    \centering
    \includegraphics[width=\linewidth]{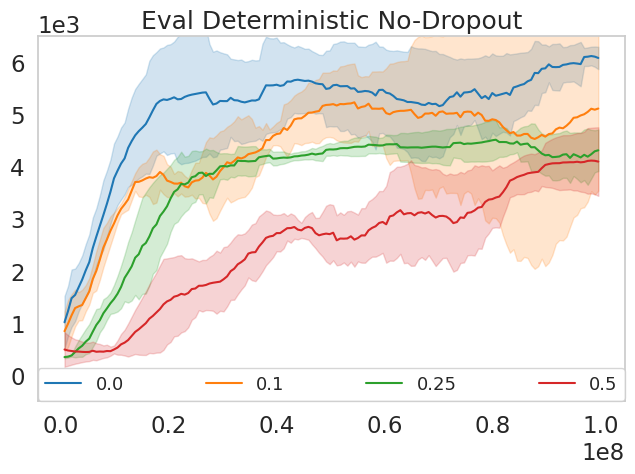}
    \subcaption[]{PPO Consistent}
\end{subfigure}
\caption{\textbf{Walker2d-v3 Deterministic Evaluation No-Dropout}: Curves are averages of three independent training runs.}
\label{fig:walker2d_eval_det_nodrop}
\end{center}
\vskip -0.2in
\end{figure*}

\clearpage
\section{Full Atari Results}
\label{sec:full_atari_results}

Hyperparameters were tuned for Beamrider then deployed without further modification on Breakout and Seaquest. The lower performance from PPO without dropout versus PPO Consistent without dropout is due to the lack of target-KL early stopping threshold for PPO. If this threshold is not disabled, PPO, due to inconsistent dropout, fails to perform any policy updates and makes no learning progress. 

\begin{figure*}[htb!]
\begin{center}
\begin{subfigure}{.24\textwidth}
    \centering
    \includegraphics[width=\linewidth]{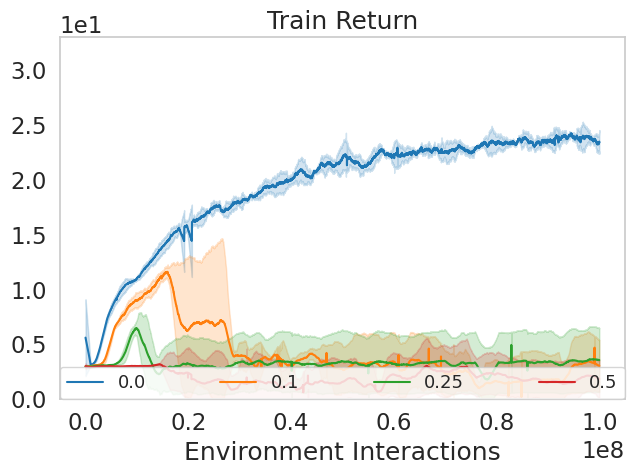}
    \subcaption[]{A2C}
\end{subfigure}
\begin{subfigure}{.24\textwidth}
    \centering
    \includegraphics[width=\linewidth]{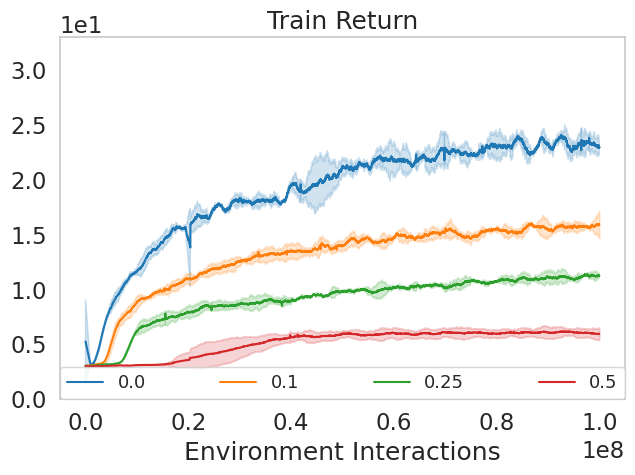}
    \subcaption[]{A2C Consistent}
\end{subfigure}
\begin{subfigure}{.24\textwidth}
    \centering
    \includegraphics[width=\linewidth]{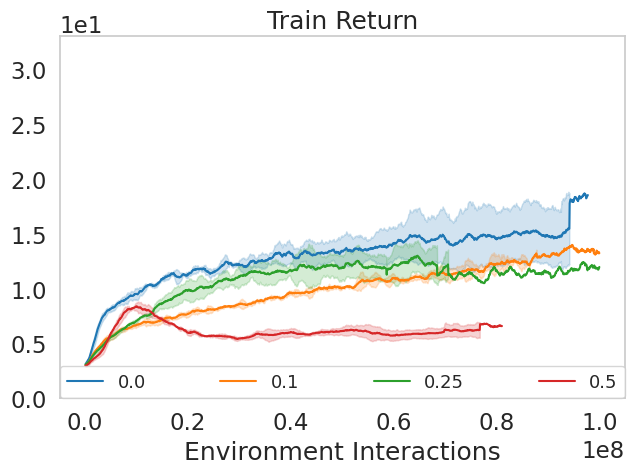}
    \subcaption[]{PPO}
\end{subfigure}
\begin{subfigure}{.24\textwidth}
    \centering
    \includegraphics[width=\linewidth]{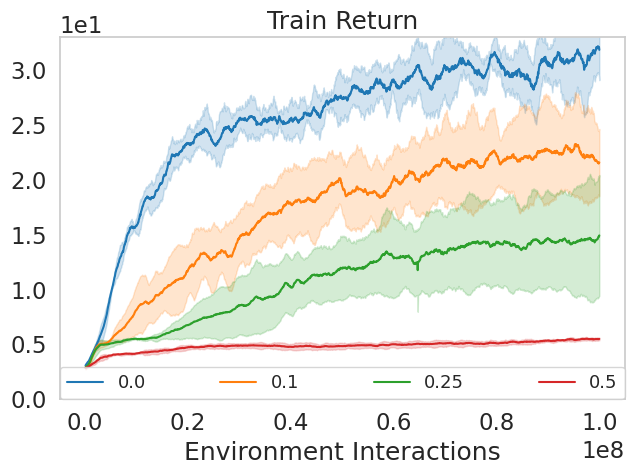}
    \subcaption[]{PPO Consistent}
\end{subfigure}
\caption{\textbf{Beamrider}: Training performance of various models on Atari Beamrider. Curves are averages of three independent training runs.}
\label{fig:beamrider_train}
\end{center}
\vskip -0.2in
\end{figure*}

\begin{figure*}[htb!]
\begin{center}
\begin{subfigure}{.24\textwidth}
    \centering
    \includegraphics[width=\linewidth]{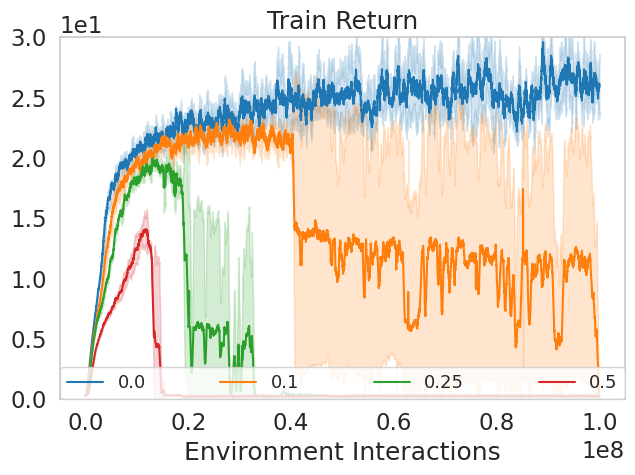}
    \subcaption[]{A2C}
\end{subfigure}
\begin{subfigure}{.24\textwidth}
    \centering
    \includegraphics[width=\linewidth]{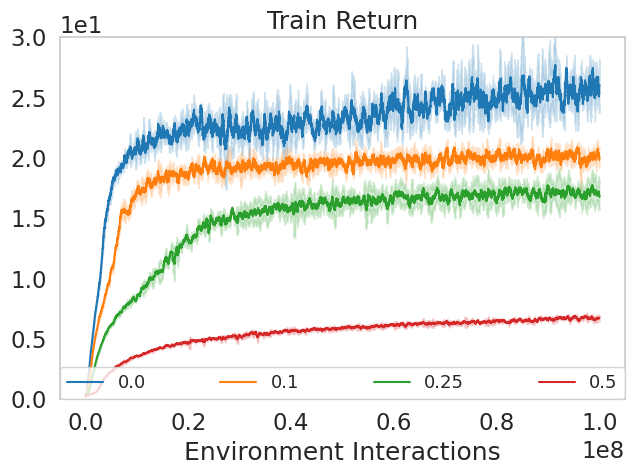}
    \subcaption[]{A2C Consistent}
\end{subfigure}
\begin{subfigure}{.24\textwidth}
    \centering
    \includegraphics[width=\linewidth]{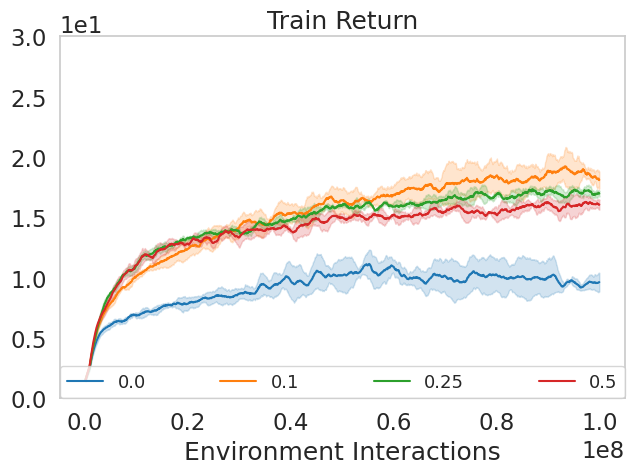}
    \subcaption[]{PPO}
\end{subfigure}
\begin{subfigure}{.24\textwidth}
    \centering
    \includegraphics[width=\linewidth]{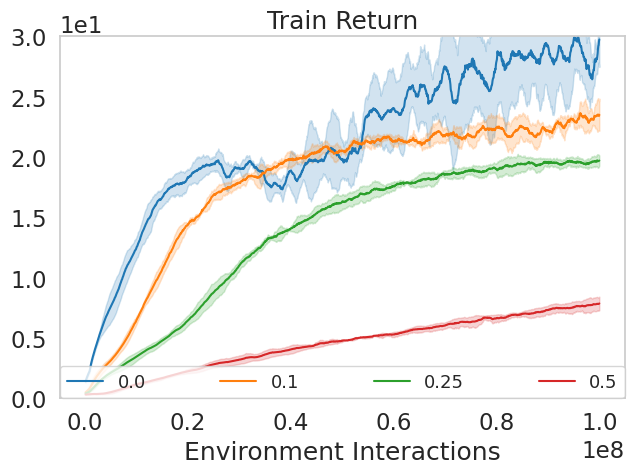}
    \subcaption[]{PPO Consistent}
\end{subfigure}
\caption{\textbf{Breakout}: Training performance of various models on Atari Breakout. Curves are averages of three independent training runs.}
\label{fig:breakout_train}
\end{center}
\vskip -0.2in
\end{figure*}

\begin{figure*}[htb!]
\begin{center}
\begin{subfigure}{.24\textwidth}
    \centering
    \includegraphics[width=\linewidth]{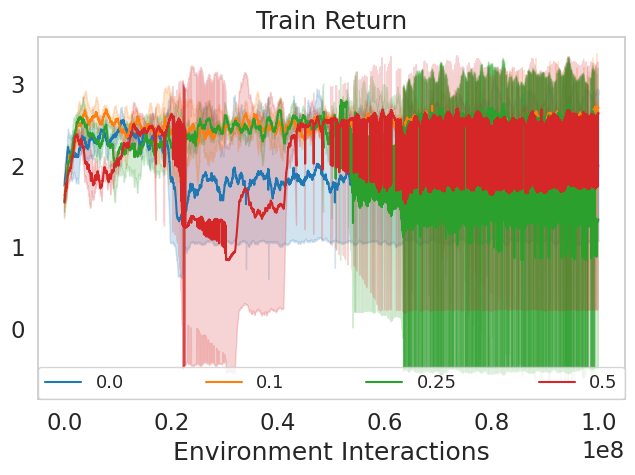}
    \subcaption[]{A2C}
\end{subfigure}
\begin{subfigure}{.24\textwidth}
    \centering
    \includegraphics[width=\linewidth]{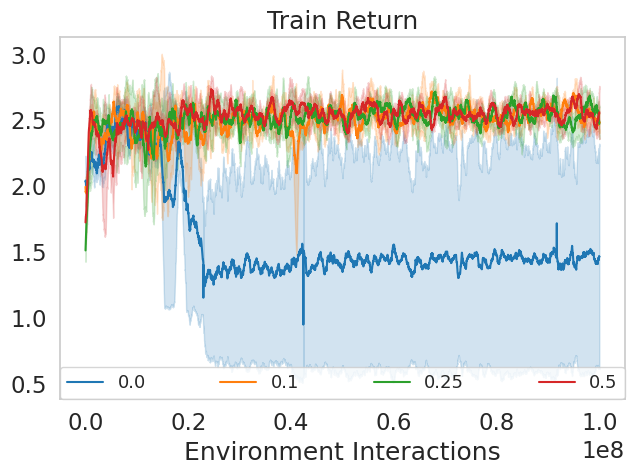}
    \subcaption[]{A2C Consistent}
\end{subfigure}
\begin{subfigure}{.24\textwidth}
    \centering
    \includegraphics[width=\linewidth]{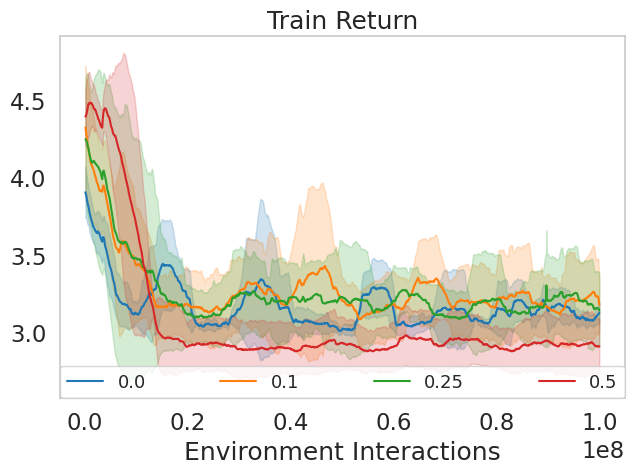}
    \subcaption[]{PPO}
\end{subfigure}
\begin{subfigure}{.24\textwidth}
    \centering
    \includegraphics[width=\linewidth]{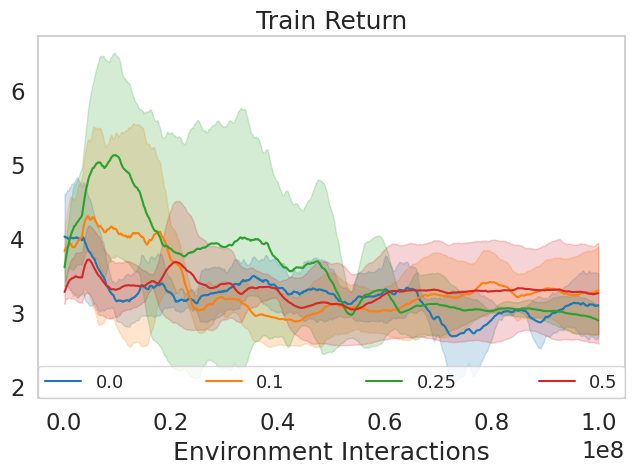}
    \subcaption[]{PPO Consistent}
\end{subfigure}
\caption{\textbf{Seaquest}: Training performance of various models on Atari Seaquest. Curves are averages of three independent training runs. Poor performance from all algorithms indicates hyperparameters which were tuned for Beamrider are suboptimal for Seaquest.}
\label{fig:seaquest_train}
\end{center}
\vskip -0.2in
\end{figure*}

\begin{figure*}[htb!]
\begin{center}
\begin{subfigure}{.24\textwidth}
    \centering
    \includegraphics[width=\linewidth]{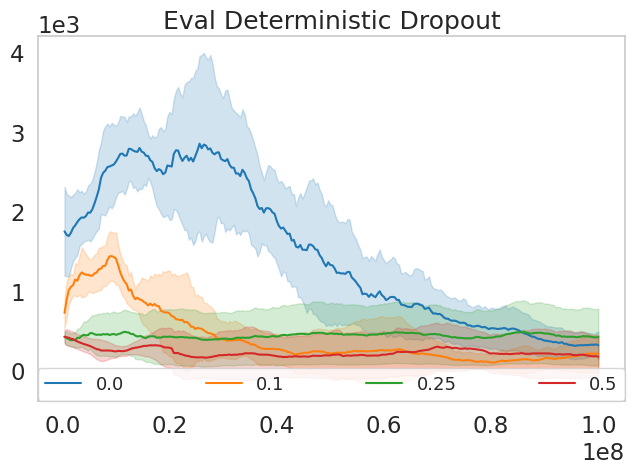}
    \subcaption[]{A2C}
\end{subfigure}
\begin{subfigure}{.24\textwidth}
    \centering
    \includegraphics[width=\linewidth]{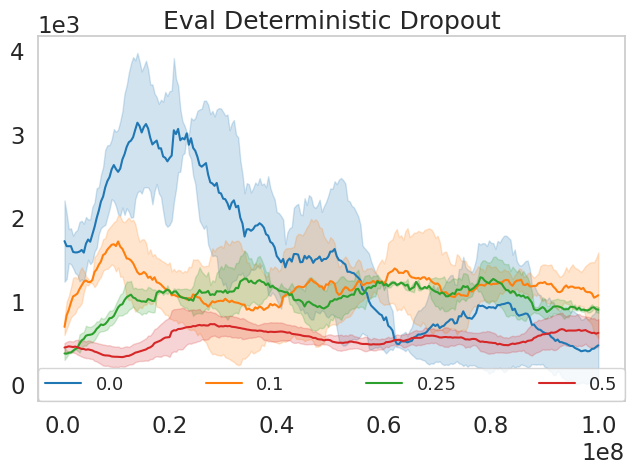}
    \subcaption[]{A2C Consistent}
\end{subfigure}
\begin{subfigure}{.24\textwidth}
    \centering
    \includegraphics[width=\linewidth]{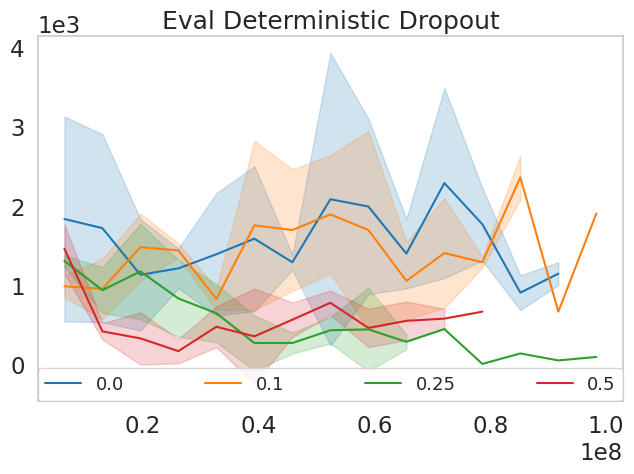}
    \subcaption[]{PPO}
\end{subfigure}
\begin{subfigure}{.24\textwidth}
    \centering
    \includegraphics[width=\linewidth]{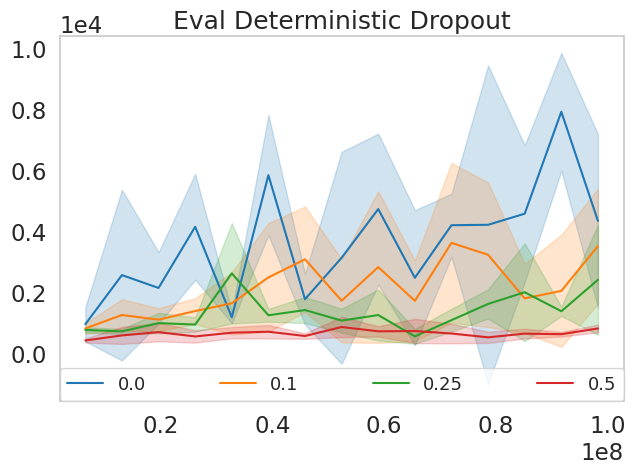}
    \subcaption[]{PPO Consistent}
\end{subfigure}
\caption{\textbf{Beamrider Deterministic Evaluation with Dropout}: Curves are averages of three independent training runs.}
\label{fig:beamrider_det_drop}
\end{center}
\vskip -0.2in
\end{figure*}

\begin{figure*}[htb!]
\begin{center}
\begin{subfigure}{.24\textwidth}
    \centering
    \includegraphics[width=\linewidth]{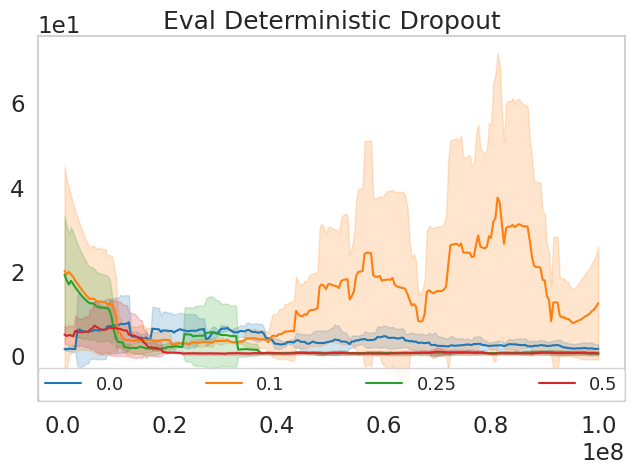}
    \subcaption[]{A2C}
\end{subfigure}
\begin{subfigure}{.24\textwidth}
    \centering
    \includegraphics[width=\linewidth]{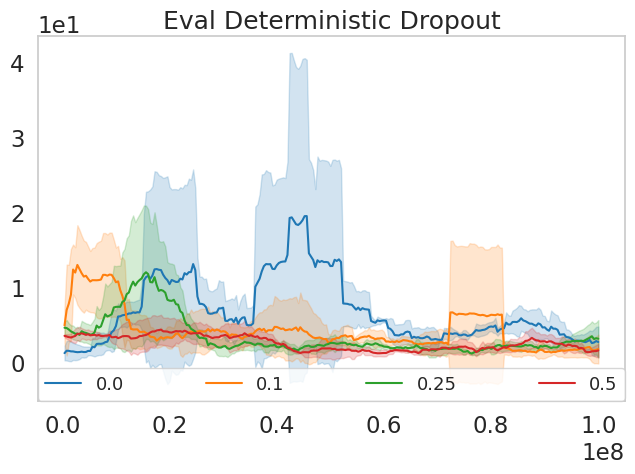}
    \subcaption[]{A2C Consistent}
\end{subfigure}
\begin{subfigure}{.24\textwidth}
    \centering
    \includegraphics[width=\linewidth]{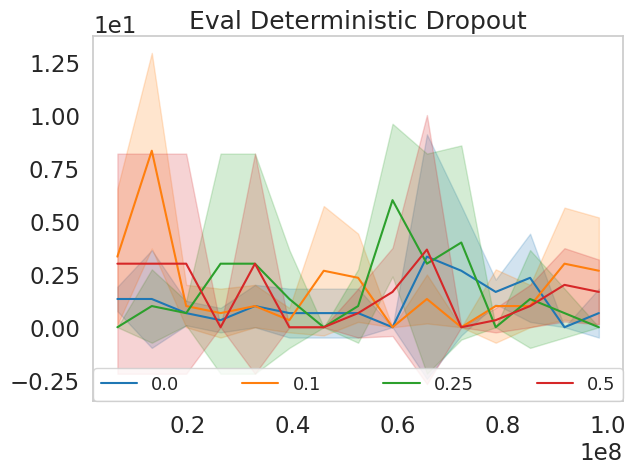}
    \subcaption[]{PPO}
\end{subfigure}
\begin{subfigure}{.24\textwidth}
    \centering
    \includegraphics[width=\linewidth]{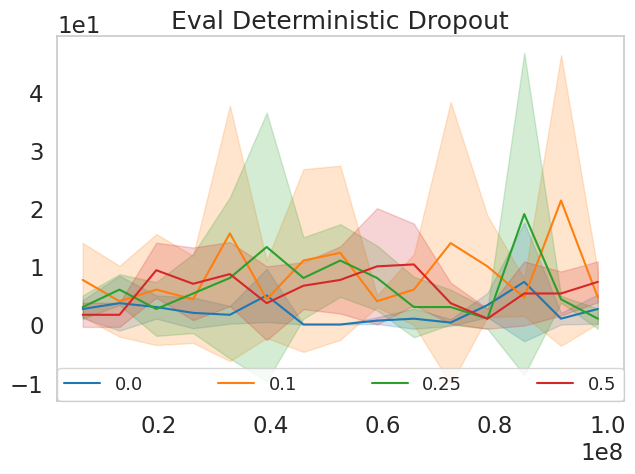}
    \subcaption[]{PPO Consistent}
\end{subfigure}
\caption{\textbf{Breakout Deterministic Evaluation with Dropout}: Curves are averages of three independent training runs.}
\label{fig:breakout_det_drop}
\end{center}
\vskip -0.2in
\end{figure*}

\begin{figure*}[htb!]
\begin{center}
\begin{subfigure}{.24\textwidth}
    \centering
    \includegraphics[width=\linewidth]{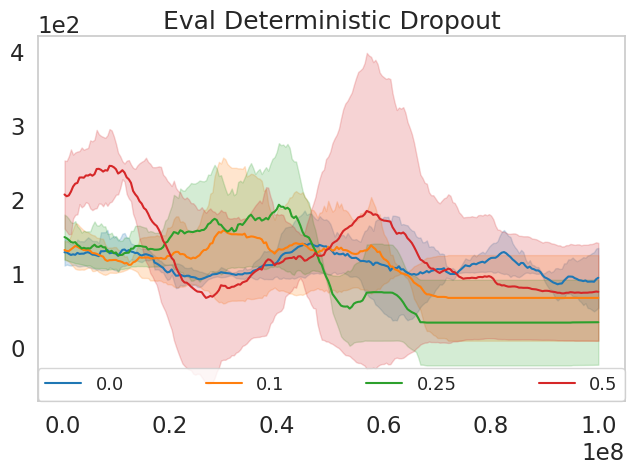}
    \subcaption[]{A2C}
\end{subfigure}
\begin{subfigure}{.24\textwidth}
    \centering
    \includegraphics[width=\linewidth]{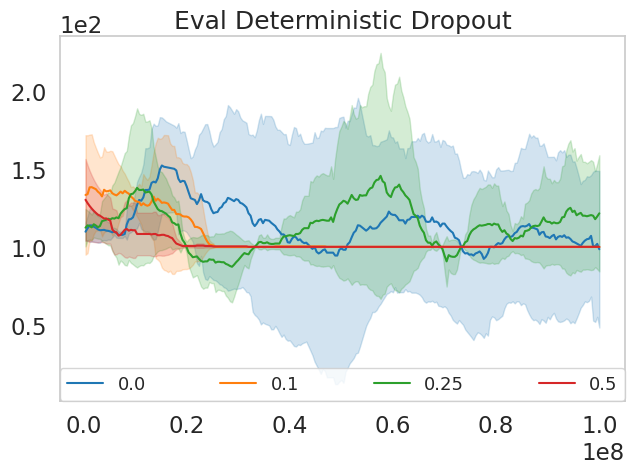}
    \subcaption[]{A2C Consistent}
\end{subfigure}
\begin{subfigure}{.24\textwidth}
    \centering
    \includegraphics[width=\linewidth]{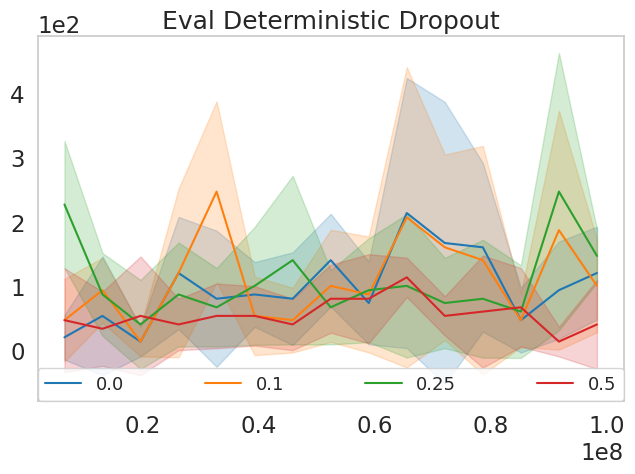}
    \subcaption[]{PPO}
\end{subfigure}
\begin{subfigure}{.24\textwidth}
    \centering
    \includegraphics[width=\linewidth]{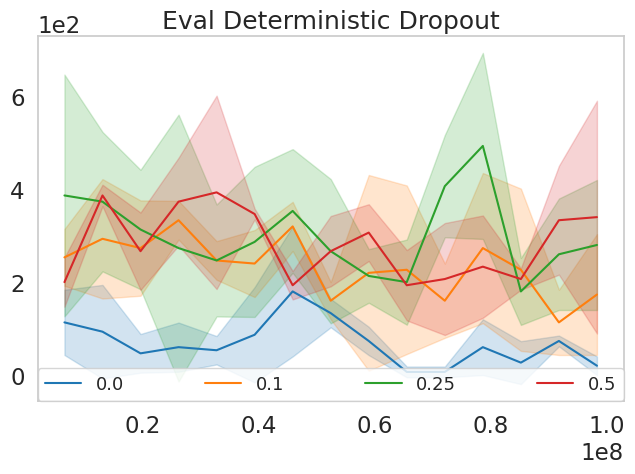}
    \subcaption[]{PPO Consistent}
\end{subfigure}
\caption{\textbf{Seaquest Deterministic Evaluation with Dropout}: Curves are averages of three independent training runs.}
\label{fig:seaquest_det_drop}
\end{center}
\vskip -0.2in
\end{figure*}

\begin{figure*}[htb!]
\begin{center}
\begin{subfigure}{.24\textwidth}
    \centering
    \includegraphics[width=\linewidth]{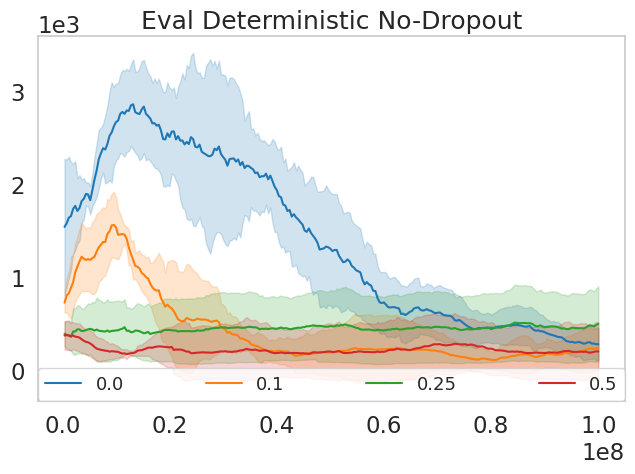}
    \subcaption[]{A2C}
\end{subfigure}
\begin{subfigure}{.24\textwidth}
    \centering
    \includegraphics[width=\linewidth]{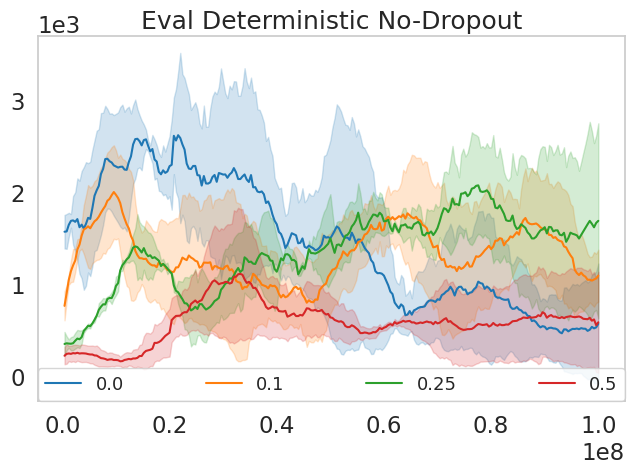}
    \subcaption[]{A2C Consistent}
\end{subfigure}
\begin{subfigure}{.24\textwidth}
    \centering
    \includegraphics[width=\linewidth]{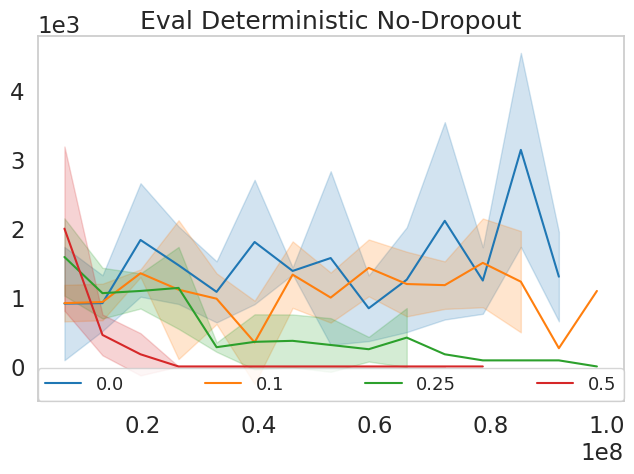}
    \subcaption[]{PPO}
\end{subfigure}
\begin{subfigure}{.24\textwidth}
    \centering
    \includegraphics[width=\linewidth]{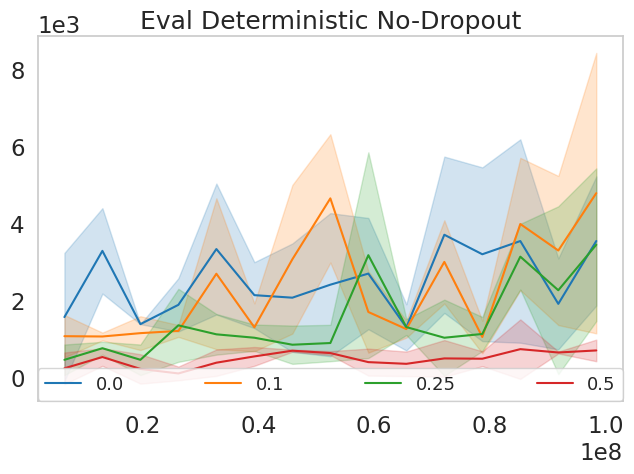}
    \subcaption[]{PPO Consistent}
\end{subfigure}
\caption{\textbf{Beamrider Deterministic Evaluation with No-Dropout}: Curves are averages of three independent training runs.}
\label{fig:beamrider_det_nodrop}
\end{center}
\vskip -0.2in
\end{figure*}

\begin{figure*}[htb!]
\begin{center}
\begin{subfigure}{.24\textwidth}
    \centering
    \includegraphics[width=\linewidth]{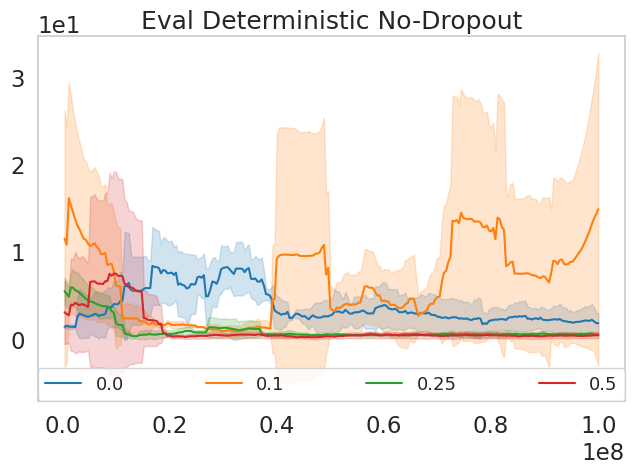}
    \subcaption[]{A2C}
\end{subfigure}
\begin{subfigure}{.24\textwidth}
    \centering
    \includegraphics[width=\linewidth]{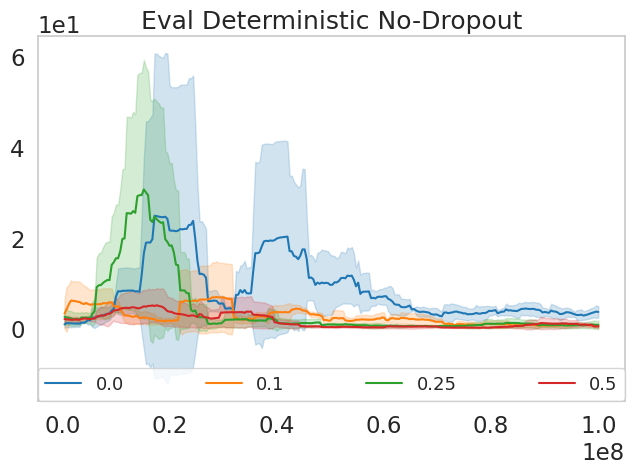}
    \subcaption[]{A2C Consistent}
\end{subfigure}
\begin{subfigure}{.24\textwidth}
    \centering
    \includegraphics[width=\linewidth]{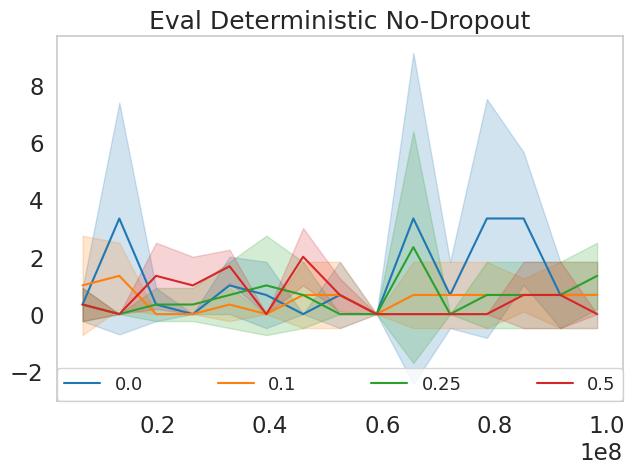}
    \subcaption[]{PPO}
\end{subfigure}
\begin{subfigure}{.24\textwidth}
    \centering
    \includegraphics[width=\linewidth]{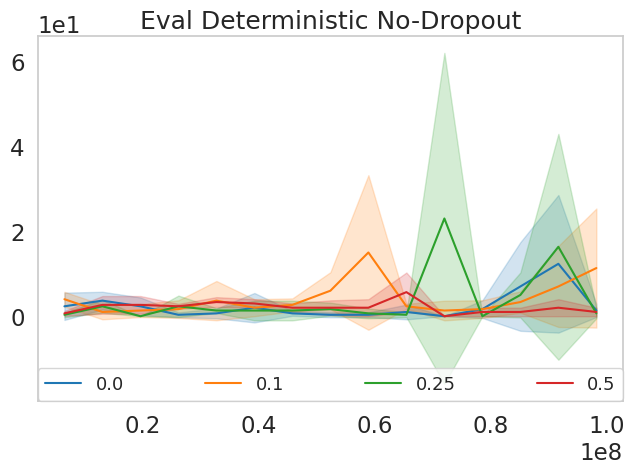}
    \subcaption[]{PPO Consistent}
\end{subfigure}
\caption{\textbf{Breakout Deterministic Evaluation with No-Dropout}: Curves are averages of three independent training runs.}
\label{fig:breakout_det_nodrop}
\end{center}
\vskip -0.2in
\end{figure*}

\begin{figure*}[htb!]
\begin{center}
\begin{subfigure}{.24\textwidth}
    \centering
    \includegraphics[width=\linewidth]{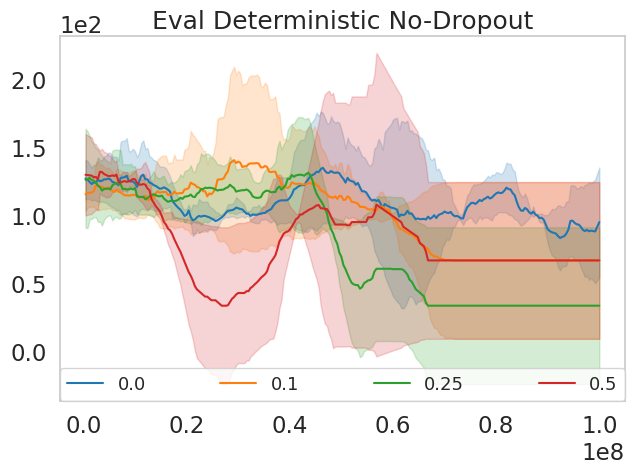}
    \subcaption[]{A2C}
\end{subfigure}
\begin{subfigure}{.24\textwidth}
    \centering
    \includegraphics[width=\linewidth]{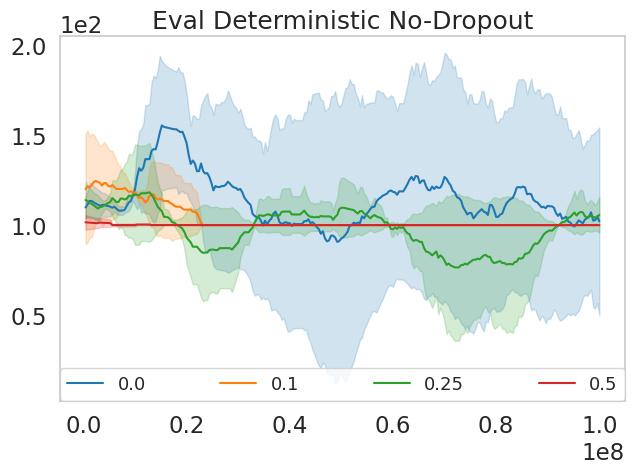}
    \subcaption[]{A2C Consistent}
\end{subfigure}
\begin{subfigure}{.24\textwidth}
    \centering
    \includegraphics[width=\linewidth]{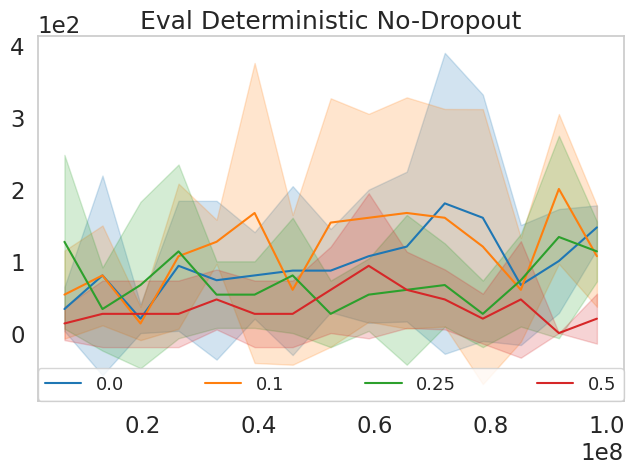}
    \subcaption[]{PPO}
\end{subfigure}
\begin{subfigure}{.24\textwidth}
    \centering
    \includegraphics[width=\linewidth]{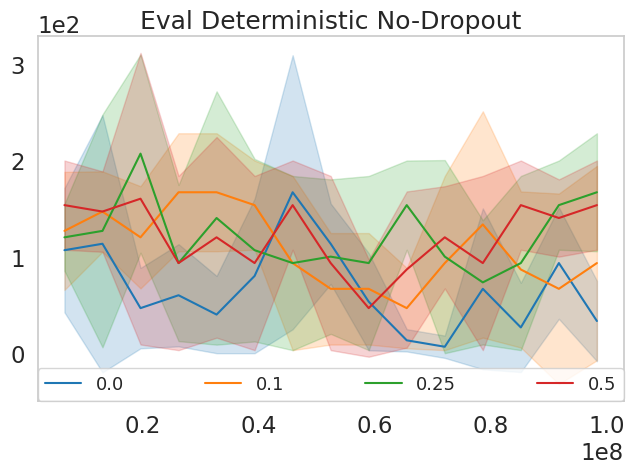}
    \subcaption[]{PPO Consistent}
\end{subfigure}
\caption{\textbf{Seaquest Deterministic Evaluation with No-Dropout}: Curves are averages of three independent training runs.}
\label{fig:seaquest_det_nodrop}
\end{center}
\vskip -0.2in
\end{figure*}


\end{document}